\documentclass[final]{cvpr}

\usepackage{times}
\usepackage{epsfig}
\usepackage{graphicx,subfigure}
\usepackage{amsmath}
\usepackage{amssymb}
\usepackage{url}
\usepackage{mathrsfs}
\usepackage{algorithmicx,algorithm}
\usepackage{verbatim}
\usepackage{amsfonts,amssymb}
\usepackage{bm}
\usepackage{mathtools}
\usepackage{makecell, multirow}
\usepackage{wrapfig}
\usepackage{picinpar}
\usepackage{array}
\usepackage{booktabs}
\usepackage{pifont}

\usepackage[pagebackref=true,breaklinks=true,colorlinks,bookmarks=false]{hyperref}

\def\cvprPaperID{3143} 
\def\confYear{CVPR 2021}

\begin{document}

\title{ORDisCo: Effective and Efficient Usage of Incremental Unlabeled Data \\ for Semi-supervised Continual Learning}

\author{Liyuan Wang\textsuperscript{\rm 1}, Kuo Yang\textsuperscript{\rm 2}, Chongxuan Li\textsuperscript{\rm 1}\thanks{Corresponding author: C. Li and J. Zhu.}
, Lanqing Hong\textsuperscript{\rm 2}, Zhenguo Li\textsuperscript{\rm 2}, Jun Zhu\textsuperscript{\rm 1*}\\
\textsuperscript{\rm 1} Dept. of Comp. Sci. \& Tech., Institute for AI, BNRist Center,\\ THBI Lab, Tsinghua University, Beijing, China\\
\textsuperscript{\rm 2} Huawei Noah's Ark Lab\\
}

\maketitle

\pagestyle{empty}
\thispagestyle{empty}

\begin{abstract}
Continual learning usually assumes the incoming data are fully labeled, which might not be applicable in real applications. In this work, we consider semi-supervised continual learning (SSCL) that incrementally learns from partially labeled data. Observing that existing continual learning methods lack the ability to continually exploit the unlabeled data, we propose deep Online Replay with Discriminator Consistency (ORDisCo) to interdependently learn a classifier with a conditional generative adversarial network (GAN), which continually passes the learned data distribution to the classifier. In particular, ORDisCo replays data sampled from the conditional generator to the classifier in an online manner, exploiting unlabeled data in a time- and storage-efficient way. Further, to explicitly overcome the catastrophic forgetting of unlabeled data, we selectively stabilize parameters of the discriminator that are important for discriminating the pairs of old unlabeled data and their pseudo-labels predicted by the classifier. We extensively evaluate ORDisCo on various semi-supervised learning benchmark datasets for SSCL, and show that ORDisCo achieves significant performance improvement on SVHN, CIFAR10 and Tiny-ImageNet, compared to strong baselines.

\end{abstract}


\section{Introduction}


Current achievements of deep neural networks (DNNs) heavily rely on large amounts of supervised data, which are expensive and difficult to acquire simultaneously. 
Therefore, the ability of continual learning (CL) on incremental training samples becomes extremely important. 
Numerous efforts have been devoted to CL, which aim to continually learn new training samples without \emph{catastrophic forgetting} of the learned data distribution \cite{parisi2019continual}. 
Existing CL methods mainly fall in two categories: weight regularization methods~\cite{parisi2019continual,kirkpatrick2017overcoming,aljundi2018memory} and replay-based methods~\cite{parisi2019continual,rebuffi2017icarl,shin2017continual}, and have achieved promising results in purely supervised settings.


In many real-world applications, nevertheless, the incremental data are often partially labeled. For example, in face recognition~\cite{roli2006semi}, a device continually obtains user data for unlocking.
These increasing data could be used to update the model for better user experience.
However, true labels of the incoming data are usually unavailable unless the user provides the password for verification.
Since frequently asking for labelling would affect user experience, most of the input data are unlabeled.
Similar scenarios occur in fingerprint identification~\cite{yuan2020semi} and video recognition~\cite{luo2017adaptive}.
Though such scenarios are common in our daily life, they are seldom studied in the CL literature. Therefore, in this paper, we focus on the challenging and realistic task that continually learns incremental partially labeled data. For simplicity, we refer to it as \emph{semi-supervised continual learning} (SSCL).

Different from supervised CL, SSCL provides insufficient supervision and a large amount of unlabeled data. As is well known, unlabeled data are crucial in semi-supervised scenarios~\cite{zhu2009introduction} but are massive to exploit. In fact, we conduct preliminary experiments in SSCL and empirically verify that the representative CL methods including the weight regularization methods~\cite{parisi2019continual,kirkpatrick2017overcoming,aljundi2018memory} and the replay-based methods~\cite{parisi2019continual,rebuffi2017icarl,shin2017continual} may not effectively exploit the unlabeled data.
Specifically, the joint training of a strong semi-supervised classifier significantly outperforms the best of existing CL strategies on the same classifier (see results in Fig.~\ref{svhn30b_MT} and Appendix A). We identify this as the \emph{catastrophic forgetting of unlabeled data} problem in SSCL.

To this end, we present deep \emph{online replay with discriminator consistency (ORDisCo)}, a framework
to continually learn a semi-supervised classifier and a conditional generative adversarial networks (GAN)~\cite{mirza2014conditional} together in SSCL. Specifically, ORDisCo is formulated as a minimax adversarial game~\cite{goodfellow2014generative}, where a generator tries to capture the underlying joint distribution of partially labeled data, and help the classifier make accurate predictions.
At the same time, the classifier also predicts pseudo-labels for unlabeled data to improve the training of the conditional generator.
In contrast to previous work~\cite{shin2017continual,wu2018memory,ostapenko2019learning}, ORDisCo replays data sampled from the conditional generator to the classifier in an \emph{online} manner, which is time- and storage-efficient to exploit the large amount of unlabeled data in SSCL. Further, to explicitly overcome the catastrophic forgetting of unlabeled data, we selectively stabilize parameters of the discriminator that are important for discriminating the pairs of old unlabeled data and their pseudo-labels predicted by the classifier.


We follow the \textit{New Instance} and \textit{New Class} scenarios of CL \cite{parisi2019continual} to split commonly used SSL benchmark datasets \cite{chongxuan2017triple,oliver2018realistic}, including SVHN, CIFAR10 and Tiny-ImageNet, as the evaluation benchmarks for SSCL. To simulate the practical scenarios, all methods continually receive a batch of data with a few labels during training. Extensive evaluations on such benchmarks show that ORDisCo can significantly outperform strong baselines in SSCL. 

In summary, our contributions include: (i) We consider a realistic yet challenging task called semi-supervised continue learning (SSCL). We provide a systematical study of existing CL strategies and show that the catastrophic forgetting of unlabeled data is the key challenge in SSCL.
(ii) We present ORDisCo to continually learn a classifier and a conditional GAN in SSCL. The generator replays data in an online manner with a consistency regularization on the discriminator to address the catastrophic forgetting of unlabeled data; (iii) We evaluate ORDisCO on various benchmarks and demonstrate that ORDisCO significantly improves both classification and conditional image generation over strong baselines in SSCL.

\section{Related Work}
\textbf{Continual learning (CL)} aims to address catastrophic forgetting in DNNs on a dynamic data distribution \cite{parisi2019continual}, which requires DNNs to learn from incoming tasks or training samples while retaining previously learned data distribution.
Current efforts in CL mainly consider the setting where large amounts of annotated data are available for training. 
Regularization-based methods selectively penalize changes of the parameters to maintain the learned data distribution, e.g., EWC \cite{kirkpatrick2017overcoming}, SI \cite{zenke2017continual} and MAS \cite{aljundi2018memory}. 
While, replay-based methods store a small memory buffer to replay representative training samples \cite{rebuffi2017icarl, castro2018end}. 
To address the imbalance between the small memory buffer and large amounts of training samples, BiC \cite{wu2019large} splits an equal number of samples from both the memory buffer and new training data as a validation set to train an additional linear layer for bias correction, while the Unified Classifier \cite{hou2019learning} regularizes the cosine similarity of features in an unsupervised fashion to normalize the prediction.
\cite{tao2020topology} maintains a topology of learned feature space, but it's difficult to learn such a CL-based topology from semi-supervised data. \cite{liu2020mnemonics} provides a more effective strategy to select the memory buffer from large amounts of labeled data, which is unavailable in SSCL.
To better recover the learned data distribution, generative replay strategies continually learn a generative model to replay generated data \cite{shin2017continual, wu2018memory, ostapenko2019learning}. However, conditional generation for CL heavily relies on large amounts of labeled data to assign correct labels. 
Thus, the extension to large amounts of unlabeled data with only a few labels is highly nontrivial. Also, the generative models are often offline saved in CL, which results in additional time and storage cost.


\textbf{CL of limited supervised data}
Few-shot continual or incremental learning (FSIL) \cite{javed2019meta,tao2020few,ayub2020brain} incrementally learns new classes from a small amount of labeled data, which generally requires a pretraining step on large amounts of base tasks and training samples for few-shot generalization. Unsupervised continual learning (UCL) \cite{rao2019continual,smith2019unsupervised} continually learns new classes from large amounts of unlabeled data. However, FSIL and UCL only make use of a small amount of labeled data or large amounts of unlabeled data, both of which are easy to acquire in many real-world applications. Also, both FSIL and UCL mainly consider learning new classes from a fixed number of training samples, rather than continually learn new instances of the learned classes. 
By contrast, our setting of SSCL aims to continually learn new instances of the learned classes and new classes from partially labeled data, which is more realistic in real-world scenarios. \cite{li2019incremental} continually learns partially labeled new instances of fixed classes through stacking generators and discriminators that significantly expand the model. While, we focus on the setting that the model should maintain a relatively constant size, which is a common assumption for continual learning. 


\textbf{Semi-supervised learning (SSL)} provides a powerful framework to leverage unlabeled data from a small amount of labels. Pseudo-Labeling (PL) \cite{lee2013pseudo} is an early work to assign the prediction of an unlabeled data that is higher than a threshold as its pseudo-label to augment the labeled data. Many more recent works of SSL on discriminative model define a regularization term to learn the distribution of unlabeled data. Consistency regularization, e.g., PI-model \cite{laine2016temporal}, makes use of the stochastic predictions of a network and adds a loss term to regularize the consistency of predictions on different passes of the same data. Mean teacher (MT) \cite{tarvainen2017mean} enforces the predictions of the classifier closer to its exponential moving average on the same batch of unlabeled data. Virtual Adversarial Training (VAT) \cite{miyato2018virtual} adds adversarial perturbations to the unlabeled data and enforces the predictions to be the same as the original one. Graph Laplacian regularization \cite{gong2015deformed} penalizes the variation of labels on the graph of manifold structure. SSL on GAN is generally built on a two-network architecture, that the discriminator is responsible for both sample quality and label prediction, including CatGAN \cite{springenberg2015unsupervised}, Improved-GAN \cite{salimans2016improved}, CGAN \cite{mirza2014conditional} and its variants \cite{radford2015unsupervised,odena2017conditional}. Triple-GAN \cite{chongxuan2017triple} applies a triple-network architecture that includes a classifier taking the role of label prediction. ISL-GAN \cite{wei2020incremental} dynamically assigns more virtual labels to unlabeled data during joint training of all the partially labeled data together. However, existing efforts of SSL on both discriminative model and GAN only consider a static data distribution rather than a dynamic one.

\section{Semi-supervised Continual Learning and Its Challenges}

In this section, we first introduce the problem formulation of continual learning (CL) of incremental semi-supervised data, i.e. semi-supervised continual learning (SSCL). Then, we provide a systematic study of existing  representative continual learning methods in SSCL. They suffer from catastrophic forgetting due to the poor usage of large amounts of unlabeled data. 

\subsection{Problem Formulation}

Compared with supervised CL, SSCL only provides a small amount of labeled data and a large amount of unlabeled data, which is key to achieve good performance. Formally, when training on a task \(t\),  SSCL is a special continual learning setting on a partially labeled dataset \(D^t = \bigcup_{b=0}^{[B]}D_{b}^t\) with \(B\) batches, where \(D_{b}^t = \{(x_{i}, y_{i})\}_{i \in b_{l}}\bigcup \{(x_{j})\}_{j \in b_{ul}} \) is a semi-supervised batch consisting of a labeled sub-batch \(b_{l}\) and an unlabeled sub-batch \(b_{ul}\). \(D_{b}^t\) is introduced when training on the current batch \(b\) and the performance on the task is evaluated after learning each batch.

The above SSCL setting is similar to continual learning of \textit{New Instance} \cite{parisi2019continual}, while the dataset \(D^t\) for SSCL only consists of small amounts of labels rather than all the labels. We consider continual learning of \textit{New Class} \cite{parisi2019continual} with partially labeled data as a natural extension, where a collection of \(T\) semi-supervised datasets \(D = \bigcup_{t=0}^{[T]}D^t\) is sequentially learned. \(D^t\) is provided during training on the task \(t\), which includes several classes. After learning each task, all of the classes ever seen are evaluated without access to the task labels, i.e. the single-head evaluation \cite{chaudhry2018riemannian}.


\subsection{A Systematic Study of Existing CL Methods}

We adapt the representative continual learning strategies, including weight regularization methods~\cite{parisi2019continual} and memory replay ones~\cite{parisi2019continual}, to SSCL by considering \emph{unlabeled data} and provides a systematical analysis. 


In particular, we conduct extensive SSCL experiments on the SVHN dataset, where we equally split the training set into 30 batches with 3 labels per class in each batch. Notably, we learn various strong semi-supervised classifiers, including MT \cite{tarvainen2017mean}, VAT \cite{miyato2018virtual}, PL \cite{lee2013pseudo} and PI \cite{laine2016temporal} models in SSCL. We consider four learning strategies for all classifiers to verify the existence of catastrophic forgetting and analyze  the underling reason, as follows: 
\begin{itemize}
  \item \textbf{Joint Training (JT):} The classifier is jointly trained on all the partially labeled data ever seen.
  \item \textbf{Sequential Training (ST):}  The classifier is sequentially trained on the incremental partially labeled data. 
  \item \textbf{Weight Regularization:} Representative approaches of weight regularization, including SI \cite{zenke2017continual}, MAS \cite{aljundi2018memory} and EWC \cite{kirkpatrick2017overcoming}, are implemented to the classifier to selectively stabilize parameters.
  \item \textbf{Memory Replay:} A memory buffer is implemented to replay old training data. Classical methods~\cite{rebuffi2017icarl,castro2018end} in CL use mean-of-feature to select samples, while in SSCL we do not have sufficient labels to perform selection. Therefore, we select data through uniform sampling, which is indeed competitive to existing selection methods as analyzed in~\cite{chaudhry2018riemannian} (See empirical results in Appendix A).
\end{itemize}

\begin{figure}[t]
    \centering
    \includegraphics[width=1\linewidth]{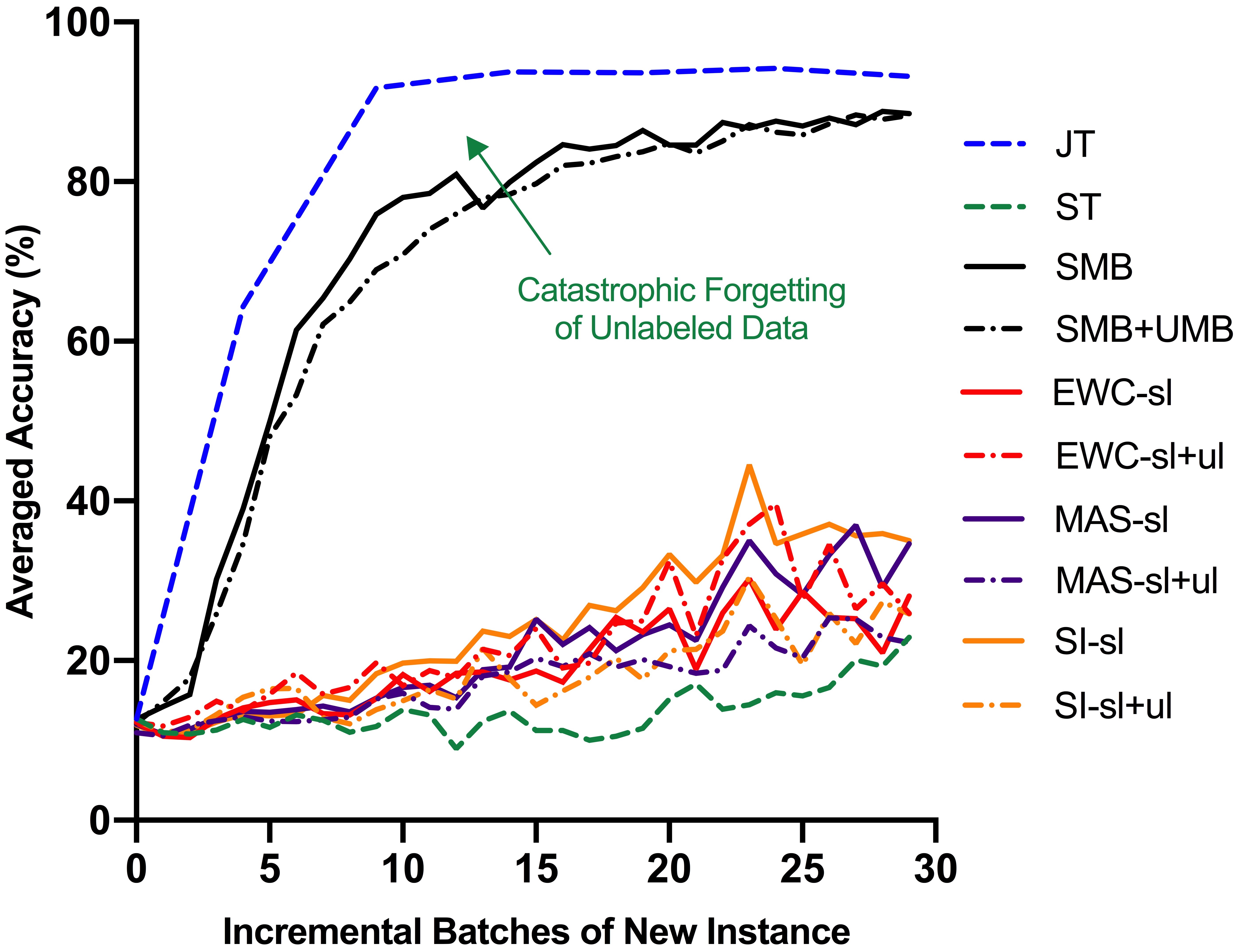}
    \caption{Baselines that combine CL strategies to address SSCL. JT: Joint training of all the training samples ever seen. ST: Sequential training on the incremental data; SMB: ST with replaying the supervised data ever seen. SMB+UMB: SMB with an unsupervised memory buffer of 10000 images to replay unlabeled data. Weight regularization methods (EWC, MAS and SI) are implemented on only the supervised loss (-sl) and both the supervised loss and the unsupervised loss (-sl+ul). The gap between JT and SMB is around \(9.70\%\) in average, which is caused by the catastrophic forgetting of unlabeled data. }
    \label{svhn30b_MT}
    \vspace{-0.2cm}
\end{figure}

We extensively search the hyperparameters of all baselines and summarize the best results. For details like the search space, we refer the readers to Appendix B.
We compare the performance of different learning strategies based on the MT classifier in Fig.~\ref{svhn30b_MT}. 
The results of other classifiers are similar and detailed in Appendix A. 

In Fig.~\ref{svhn30b_MT}, it can be seen that
ST on the incremental partially labeled data significantly underperforms JT, suggesting the existence of catastrophic forgetting. 
Besides, simply adopting weight regularization methods~\cite{parisi2019continual,kirkpatrick2017overcoming,aljundi2018memory,zenke2017continual} on a semi-supervised classifier cannot address the issue in SSCL. We hypothesize that the regularization of parameter changes in such methods limits the learning of incremental data, see more experiments and detailed discussion in Appendix B. A potential future work can be designing proper regularization terms in SSCL.

In comparison, using a supervised memory buffer (SMB), whose maximum size is of 900 images, to replay labeled data ~\cite{parisi2019continual,chaudhry2018riemannian} can effectively reduce the gap from JT but still has a large space to improve.
Further, we evaluate a strategy with an additional unsupervised memory buffer (SMB+UMB), whose maximum size is of 10,000 images, to replay unlabeled data. Indeed, the extra buffer is about twice the size of a typical generator used in our method. Given an extra large memory buffer for unlabeled data, however, we observed that the performance of the memory replay approach cannot be further improved, implying that the unlabeled data are not fully exploited. In fact, we visualize the coverage of the UMB on the distribution of all the training data using t-SNE~\cite{maaten2008visualizing} embedding (see details in Appendix A). As shown in Fig.~\ref{coverage} (a), although the UMB has already stored around \(10 \%\) data of the entire dataset, it still cannot cover the distribution of training data very well.

Therefore, a common issue of existing CL strategies in the semi-supervised scenarios is that they cannot effectively capture and exploit the distribution of unlabeled data. We identify this as the \emph{catastrophic forgetting of unlabeled data} problem, which is challenging because the incremental unlabeled data are massive and lack annotations. This motivates us to propose two new strategies to continually learn a generative model that can successfully capture the data distribution (see Fig.~\ref{coverage}, b) given partially labeled data.

\begin{figure}
  \centering
  \subfigure[UMB]{
  \includegraphics[width=1.5in]{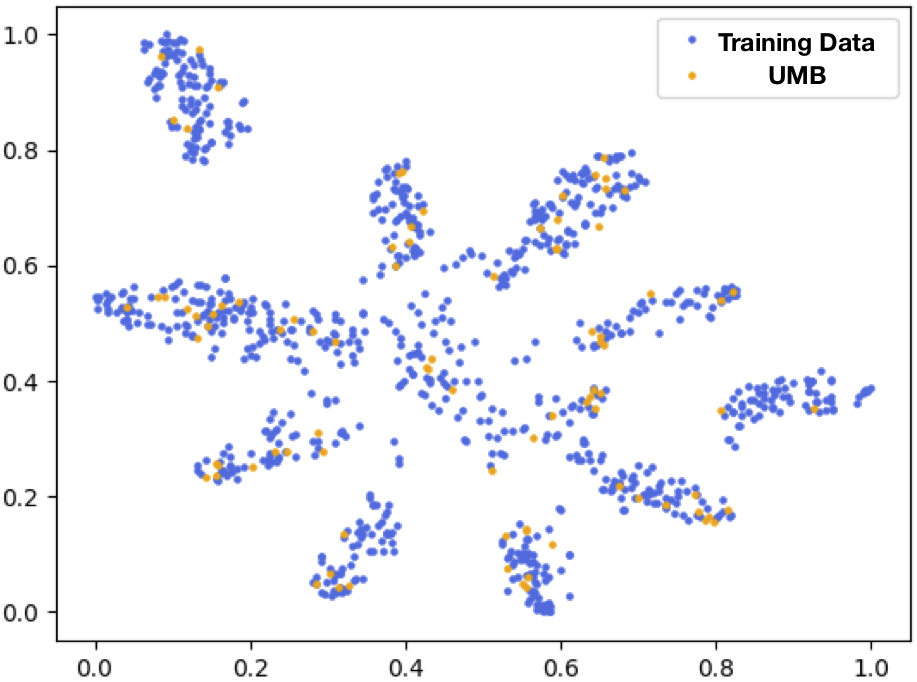}}
  \hspace{0.01in}
  \subfigure[ORDisCo]{
    \includegraphics[width=1.5in]{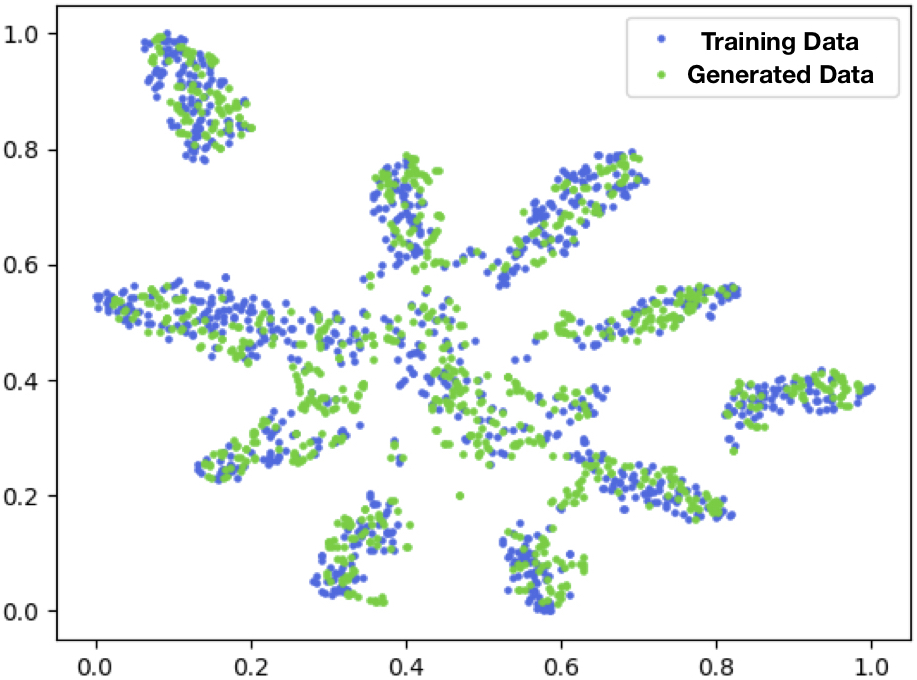}}
  \caption{Data coverage of an unsupervised  memory buffer, i.e. UMB (a), and the samples generated from ORDisCo (b). ORDisCo has a less storage cost but achieves a better result.}
  \label{coverage}
  \vspace{-0.3cm}
\end{figure}


\section{Method}

Based on the empirical analysis above, in this section, we first describe our framework that interdependently trains a classifier, a discriminator and a generator on the incremental semi-supervised data in Sec.4.1. Then, in Sec.4.2, we present two new strategies to mitigate catastrophic forgetting of the unlabeled data.

\begin{figure}[t]
    \centering
    \includegraphics[width=0.90\linewidth]{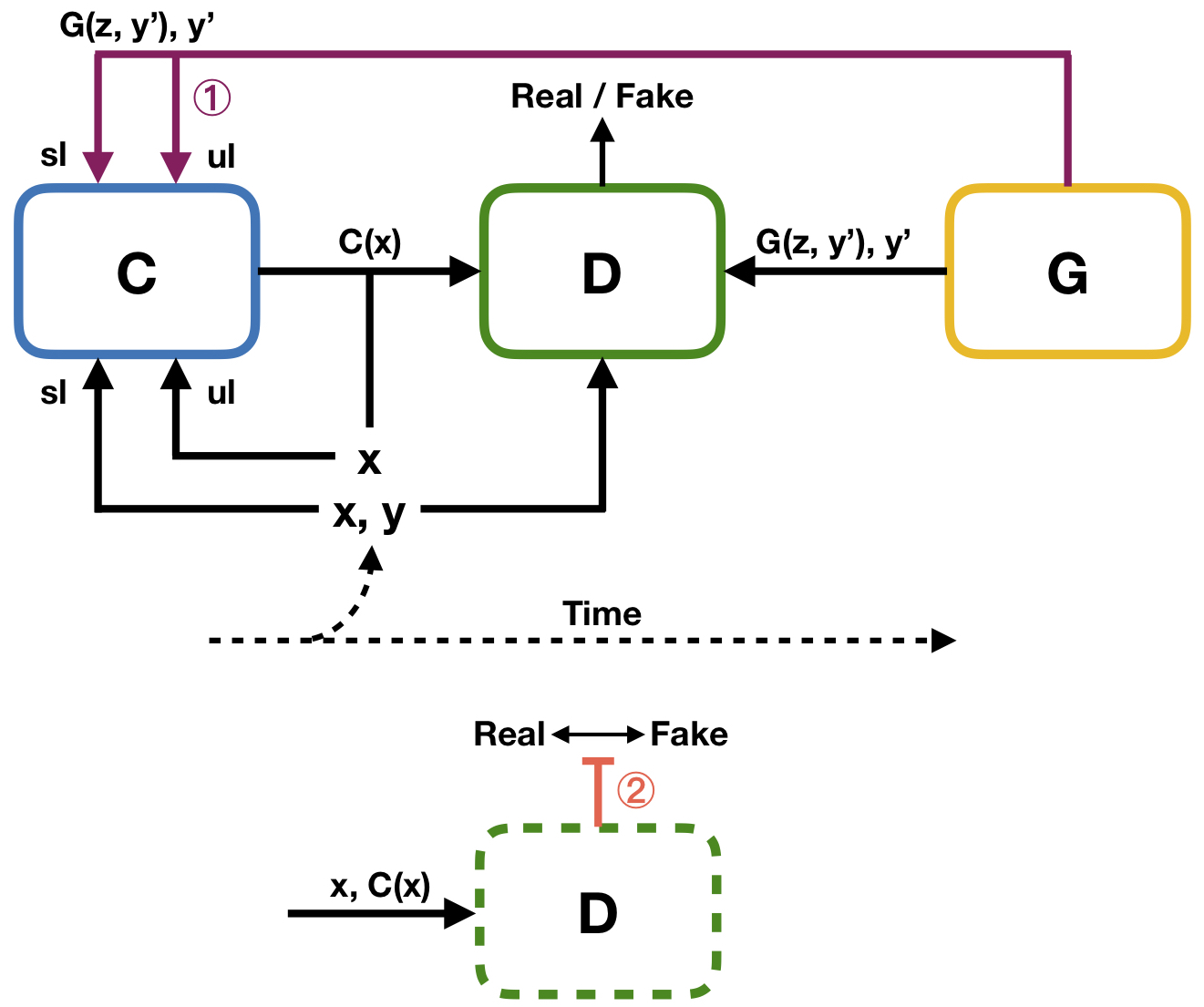}
    \caption{Deep online replay with discriminator consistency (ORDisCo). The partially labeled data are incrementally learned in the interplay of \(C\), \(G\), and \(D\). To mitigate catastrophic forgetting and better exploit the unlabeled data, we implement \ding{172} online semi-supervised generative replay in \(C\) and \ding{173} stabilization of discrimination consistency in \(D\). \(z\) and \(y'\) are the noise and the labels for conditional sampling, where the generated samples are applied as both labeled and unlabeled data for replay. Here, \(sl\) and \(ul\) stand for supervised and unsupervised losses, respectively. }
    \label{DisCo}
    \vspace{-0.3cm}
\end{figure}

\subsection{Conditional Generation on Incremental Semi-supervised Data}

Different from existing generative replay methods~\cite{shin2017continual,wu2018memory,ostapenko2019learning}, we consider a SSCL setting where partially labeled data are given. 
Inspired by the state-of-the-art semi-supervised learning (SSL) GAN~\cite{chongxuan2017triple,li2019triple}, we build up a triple-network structure that continually learns a classifier together with a conditional generative adversarial network (GAN) to better exploit the unlabeled distribution from a few labeled data in SSCL.
Both the classifier and the conditional generator produce ``fake'' data-label pairs and the discriminator focuses on discriminating if a data-label pair is real or fake. Compared to other SSL GANs~\cite{springenberg2015unsupervised,salimans2016improved,mirza2014conditional}, our framework is theoretically optimal under a nonparametric assumption and ensures that the GAN to capture the data distribution and the classifier to make predictions actually help each other~\cite{chongxuan2017triple,li2019triple}.

Formally, the entire model includes a classifier \(C\), a generator \(G\) and a discriminator \(D\). Correspondingly, the optimization problem is to learn three groups of parameters as \(\{ \theta_C, \theta_G, \theta_D\}\). In such a triple-network architecture, \(C\) tries to improve its ability of classification and generate pseudo-labels for the conditional generation in \(D\) and \(G\). While, \(D\) and \(G\) continually learn and recover the data distributions seen in the incremental semi-supervised data. The general framework is shown in Fig.~\ref{DisCo} and we will describe the objective function for each component below.

To learn a classifier \(C\) from partially labeled data, the loss functions can be generally defined as a supervised loss \(L_{sl} (\theta_{ C})\) and an unsupervised loss \(L_{ul} (\theta_{ C})\). Since our motivation is to continually capture and exploit the unlabeled data, we build up our method with a supervised memory buffer \(smb\) to replay a few labeled data. We define \(L_{sl} (\theta_{ C})\) and \(L_{ul} (\theta_{ C})\) as a cross entropy term and a consistency regularization term~\cite{tarvainen2017mean}:
\begin{equation}
{L}_{C,pl}(\theta_{ C}) = L_{sl} (\theta_{ C})+ L_{ul} (\theta_{ C}),
\label{eq:pl loss}
\end{equation}
\begin{equation}
L_{sl} (\theta_{ C}) = \mathbb{E}_{x,\, y\sim b_l \cup smb}\,[-y \, \log\,C(x)],
\label{eq:sl loss}
\end{equation}
\begin{equation}
L_{ul} (\theta_{ C}) = \mathbb{E}_{x\sim b_{ul}, \epsilon, \epsilon'}\,[\parallel C(x, \epsilon) - C'(x, \epsilon') \parallel^2],
\label{eq:ul loss}
\end{equation}
where \(C'\) is the exponential moving average of \(C\) \cite{tarvainen2017mean}, \(\parallel \cdot \parallel^2\) is the square of the \(l_2\)-norm, and \(\epsilon\) and \(\epsilon'\) denote two different random noises, e.g. dropout masks. We mention that there are other options of unlabeled losses for SSL such as~\cite{miyato2018virtual,laine2016temporal,sohn2020fixmatch}. We leave a systematical study of this for future work.

Next, we design a semi-supervised discriminator, which aims to distinguish if a data-label pair is from the labeled dataset or is a fake one:
\begin{equation}
\begin{split}
{L}_{D,pl}(\theta_{ D}) = 
& \,\, \mathbb{E}_{x, \, y\sim b_l \cup smb}\,[\log ( D(x, y) )] \\
& + \alpha \, \mathbb{E}_{y'\sim p_{y'},  z \sim p_{z}}\,[\log( 1- D(G(z, y'), y') )] \\
& + (1-\alpha) \, \mathbb{E}_{x\sim bu}\,[\log(1 - D(x, C(x)) )],
\end{split}
\label{eq:netD loss}
\end{equation}
where the sampling distribution of label \(y'\) is uniform \({p}_{y'} = \textit{U}\{ 1,2, ..., k\}\) among all the \(k\) classes ever seen and the noise \(z\) is Gaussian \(p_{z} = N(0, 1)\). The first two terms are the regular loss functions for conditional GAN, computed on the labeled training data and the generated data, respectively. While the third term is computed on the unlabeled sub-batch \(b_{ul}\) with the pseudo-labels predicted by \(C\). We assign \(\alpha\) and \(1-\alpha\) as the coefficients of the last two terms to keep balance with the first term.

Finally, the generator applies a regular loss function for conditional generation as follows:
\begin{equation}
{L}_{G}(\theta_{G}) = \mathbb{E}_{y'\sim p_{y'},  z \sim p_{z}}\,[-\log(1 - D(G(z, y'), y'))]. \\
\label{eq:netG loss}
\end{equation}

\subsection{Improving Continual Learning of Unlabeled Data}
\begin{figure}[t]
    \centering
    \includegraphics[width=1\linewidth]{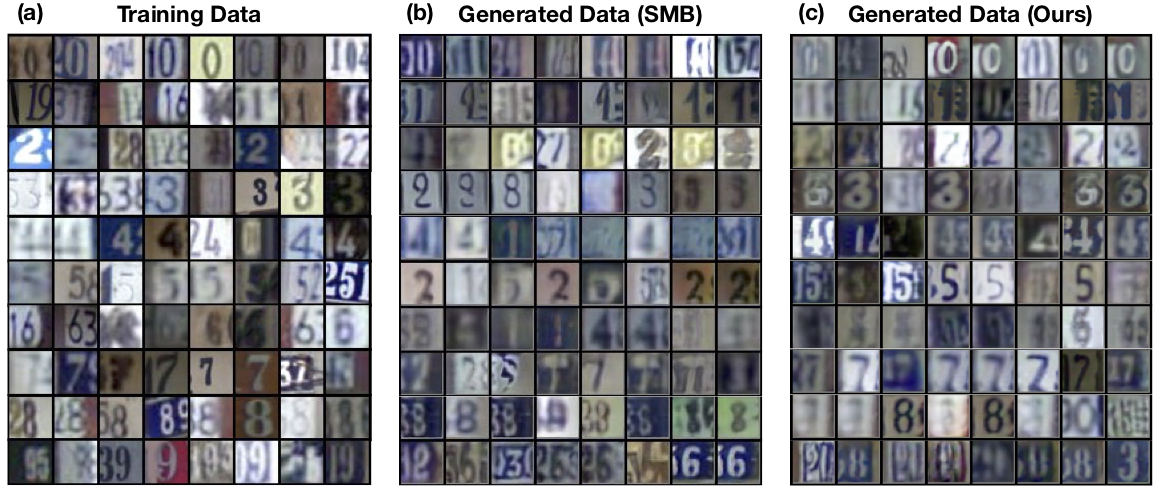}
    \caption{Conditional samples in SSCL on SVHN (after learning 5 batches of partially labeled data). (a) Training data in SVHN. (b) Conditional samples from a conditional GAN baseline with replay of supervised data (SMB). (c) Conditional samples from ORDisCo with the same SMB. Each row has a fixed label. Our method outperforms the baseline in terms of label consistency because of the two strategies proposed.}
    \label{T+G_data}
    \vspace{-0.2cm}
\end{figure}

In such a framework, after training on a semi-supervised batch \(b_i\), \(C\) learns the distribution of \(b_{i,ul}\) from \(b_{i,l}\) for classification while \(G\) and \(D\) learns that for conditional generation. When accommodating for the next batch \(b_{i+1} \), all of the three networks adjust the learned parameters of the previous batch and suffer from \emph{catastrophic forgetting}, especially on unlabeled data. Because unlabeled data are key to solve semi-supervised tasks, without effective strategies to continually learn the unlabeled data from limited labels, the generative model would fail to assign correct labels to the conditional samples during the incremental learning (see Fig.~\ref{T+G_data} for empirical evidence). To this end, we propose two strategies: online semi-supervised generative replay and stabilization of discrimination consistency.

\begin{table}[t]
	\centering
	\caption{Time and storage complexity of different generative replay strategies. \(B\) is the total number of batches to learn. 
}\smallskip
	\resizebox{1\columnwidth}{!}{
	\begin{tabular}{cccc}
		\specialrule{0.01em}{1.2pt}{1.5pt}
            &Strategy (1) \cite{ostapenko2019learning} &Strategy (2) \cite{shin2017continual,wu2018memory} &ORDisCo \\
          \specialrule{0.01em}{1.2pt}{1.5pt}
           Algorithm & offline & offline & online\\
           Time &O(\(B\)) &O(\(B^2\))  &O(\(B\)) \\
           Storage &O(\(B\)) &O(\(1\)) & O(\(1\)) \\
      \specialrule{0.01em}{1.5pt}{1.5pt}
	\end{tabular}
	}
	\label{complexity}
	\vspace{-.4cm}
 \end{table}

First, we propose a time- and storage-efficient strategy of online semi-supervised generative replay to improve CL of unlabeled data in \(C\). 
Existing generative replay methods \cite{shin2017continual,wu2018memory,ostapenko2019learning,xiang2019incremental,liu2020generative} often offline sample generated data or embedding, or save the old generators, which results in additional time and storage cost.
Let's consider two main strategies for offline generative replay: (1) All old generators learned on each task or batch are saved, and replay conditional samples to a classifier \cite{ostapenko2019learning}; and (2) After learning each task or batch, the generator is saved. In the next task or batch, the generator replays conditional samples with the new training samples to update the generator and the classifier \cite{shin2017continual,wu2018memory, xiang2019incremental,liu2020generative}.
In contrast, we online generate conditional samples from \(G\) and replay them to \(C\) rather than offline save a model for replay. As shown in Table \ref{complexity}, our online strategy is more time- and storage-efficient to exploit the large amounts of unlabeled data in SSCL (See detailed calculation in Appendix E). The conditional samples are used as both labeled data and unlabeled data to fully exploit the learned data distribution:
\begin{equation}
{L}_{C}(\theta_{C}) = {L}_{C,pl}(\theta_{ C}) + L_{G\rightarrow C} (\theta_{C}),
\label{eq:netC loss}
\end{equation}
where the first term is defined in Eqn.~\ref{eq:pl loss} and the second term exploits the data sampled from the generator in an online manner, as follows: 
\begin{equation}
\begin{split}
& L_{G\rightarrow C} (\theta_{C}) =
 \mathbb{E}_{y'\sim p_{y'},  z \sim p_{z}}\,[-y' \, \log\,C(G(z, y'))] \\
& + \mathbb{E}_{y'\sim p_{y'},  z \sim p_{z}, \epsilon, \epsilon'}\,[\parallel C(G(z, y'), \epsilon) - C'(G(z, y'), \epsilon') \parallel^2],
\end{split}
\label{eq:replay loss}
\end{equation}
which is in analogy to the losses of labeled and unlabeled training data in Eqn.~\ref{eq:sl loss}, \ref{eq:ul loss}.

Further, to explicitly mitigate catastrophic forgetting of unlabeled data in \(D\), we add a regularization term to selectively penalize changes of the parameters \(\theta_D\) from the old parameters \(\theta_D^*\) learned from the previous batches \(1:b\), depending on their contributions to the consistency of discriminating data-label pairs:

\begin{equation}
{L}_{D}(\theta_{ D}) = {L}_{D,pl}(\theta_{ D}) + \lambda \sum_{i} {\xi}_{1:b, i}( \theta_{D,i } - \theta_{D, i}^{*})^{2},
\label{eq:netD+reg loss}
\end{equation}
where the first term is defined in Eqn.~\ref{eq:netD loss} and \({\xi}_{1:b, i}\) is the strength of penalty on the changes of parameter \(i\). It is computed as 
\begin{equation}
 {\xi}_{1:b, i} =  ((b-1) \times {\xi}_{1:b-1, i} +  {\xi}_{b, i}) / b,
\end{equation}
where 
\begin{equation}
 {\xi}_{b, i} = \mathbb{E}_{x\sim bu} \, \left[ \parallel \frac{\partial \parallel (D(x, C(x))\parallel^2}{\partial \theta_{D,i}} \parallel \right].
\end{equation}

 After training on each batch \(b\), we measure \({\xi}_{b, i}\) through the expected norm of gradients on each parameters to the squared \(l_2\)-norm of discriminating the pairs of unlabeled data and their pseudo-labels predicted by \(C\). Then we update \({\xi}_{1:b, i}\) by its average. Therefore, the changes of parameters will be penalized if they change the predictions of data-label pairs on the old unlabeled batches and the learned pseudo-labels.
Note that, since the regularization is in a re-weighted weight decay form, which is non-negative and is zero in expectation given an optimal discriminator, the regularization won't hurt the consistency and convergence given different initialization and data sequences.


\section{Experiment}
In this section, we first introduce the benchmark dataset and experiment settings of SSCL. Then we show the results of our method and the baselines on SSCL. 

\subsection{Experiment Setup}
\textbf{Dataset:}
SVHN \cite{netzer2011reading} is a dataset of colored digits, including 73,257 training samples and 26,032 testing samples of size 32 $ \times $ 32. CIFAR10 \cite{krizhevsky2009learning} is a dataset of 10-class colored images with 50,000 training samples and 10,000 testing samples of size 32 $ \times $ 32. Tiny-ImageNet is a dataset of 200-class natural images with 500, 50, 50 samples per class for training, validating and testing, respectively. We randomly choose 10 or 20 classes of images from Tiny-ImageNet, resized to 32 $ \times $ 32 or 64 $ \times $ 64. For the experiment of \textit{New Instance}, we randomly split the training images of SVHN, CIFAR10 and Tiny-ImageNet into 30, 30 and 10 batches. Then we allocate a small amount of labels to each batch as the benchmark of SSCL. To simplify the notation, the benchmarks are denoted as ``dataset-(number of labels / class / batch)'', including SVHN-1, SVHN-3, CIFAR10-5, CIFAR10-13 and Tiny-ImageNet-5.
A model incrementally learns the semi-supervised batch and is evaluated on the test set after training on each batch.
For the experiment of \textit{New Class}, we use the same semi-supervised splits as \textit{New Instance}, but construct the sequence as 5 binary classification tasks, where the partially labeled data of two new classes are provided in each task. We follow the single-head evaluation \cite{chaudhry2018riemannian} that all of the classes ever seen are evaluated without access to the task labels.

\textbf{Architecture:}
In our preliminary experiments (Fig. \ref{svhn30b_MT}, Appendix A), the classifiers use a Wide ResNet architecture (WRN-28-2), which is widely applied in SSL \cite{oliver2018realistic}. To accelerate training, ORDisCo applies a much simplified classifier similar to \cite{li2019triple}, including 9 convolution layers without residual connections. The simplified classifier achieves a comparable performance on the incremental semi-supervised data as WRN-28-2 (Fig. \ref{svhn30b_MT}, Appendix A). Our generator and discriminator use a similar architecture as \cite{li2019triple}, with spectral norm to stabilize training. We keep ORDisCo and all the baselines that directly compare with ORDisCo using the same architecture for fair comparison. Please see Appendix G for hyperparameters of ORDisCo.

\textbf{Baselines:} 
All the SSL methods in our preliminary experiment (Fig. \ref{svhn30b_MT}, Appendix A) follow the same implementation as \cite{oliver2018realistic}, validated in Appendix B. Because mean teacher (MT) \cite{tarvainen2017mean} can be easily extended to many recent works \cite{iscen2019label,athiwaratkun2018there,luo2018smooth}, we choose it as the base model of SSL classifier in ORDisCo. We compare ORDisCo with the strongest baseline in the preliminary experiment, that the classifier is accessible to the labeled data ever seen through a supervised memory buffer (SMB). Since our motivation is to effectively capture and exploit the unlabeled data, we consider two extensions of SMB: (1) We combine SMB with an unsupervised memory buffer (UMB) of a similar size as our generator; and (2) We implement the \textit{Unified Classifier} \cite{hou2019learning}, the state-of-the-art method of incremental learning with memory replay, to better exploit the unsupervised memory buffer (UMB+UC). The Unified Classifier aims to address the imbalance between the small memory buffer and large amounts of training samples through regularizing the cosine distance of feature extractors in an unsupervised fashion to unify the prediction. 

\newcommand{\tabincell}[2]{\begin{tabular}{@{}#1@{}}#2\end{tabular}}

\begin{table}[t]
	\centering
	\caption{Averaged accuracy (\%)  of SSCL on Tiny-ImageNet. We randomly split training images of Tiny-ImageNet into 10 batches with equal number of labels (5 labels / class) in each batch. Here we show the results on the earlier 5 batches.}\smallskip
	\resizebox{1.00\columnwidth}{!}{
	\begin{tabular}{clccccc}
		\specialrule{0.01em}{1.2pt}{1.5pt}
		\multicolumn{1}{c}{}  & \multicolumn{1}{c}{Method}  & \multicolumn{1}{c}{Batch 1} & \multicolumn{1}{c}{Batch 2}  & \multicolumn{1}{c}{Batch 3} & \multicolumn{1}{c}{Batch 4} & \multicolumn{1}{c}{Batch 5}    \\
		\specialrule{0.01em}{1.5pt}{1.5pt}
		\multirow{3}*{\tabincell{c}{10-class \\ \(32 \times 32\)}}
        &SMB&43.00 &46.60 &55.40 &63.00 &59.80 \\
        &SMB+UMB&43.80 &47.60 &55.20 &62.20 &63.40 \\
       &ORDisCo&\textbf{45.20} &\textbf{54.40} &\textbf{59.60} &\textbf{65.40} &\textbf{69.20} \\
      \specialrule{0.01em}{1.5pt}{1.5pt}
      		\multirow{3}*{\tabincell{c}{20-class \\ \(32 \times 32\)}}
       &SMB&24.90 &32.50 &32.80 &38.70 &42.30 \\
       &SMB+UMB&\textbf{25.20} &33.10 &33.60 &37.70 &42.60 \\
       &ORDisCo&\textbf{25.20} &\textbf{33.30} &\textbf{36.50} &\textbf{42.00} &\textbf{45.10} \\
      \specialrule{0.01em}{1.5pt}{1.5pt}
      	\multirow{3}*{\tabincell{c}{10-class\\\(64 \times 64\)}}
        &SMB& \textbf{52.00} &53.00 &58.60 &66.00 &68.40  \\
        &SMB+UMB&51.60 &57.00 &62.80 &68.40 &68.80 \\
        &ORDisCo&51.20 &\textbf{60.20} &\textbf{65.20} &\textbf{70.40} &\textbf{72.80}\\
      \specialrule{0.01em}{1.5pt}{1.5pt}

	\end{tabular}
	}
	\label{tiny}
	\vspace{-.2cm}
 \end{table}

\subsection{SSCL of New Instance}

\begin{figure}[t]
    \centering
    \includegraphics[width=0.80\linewidth]{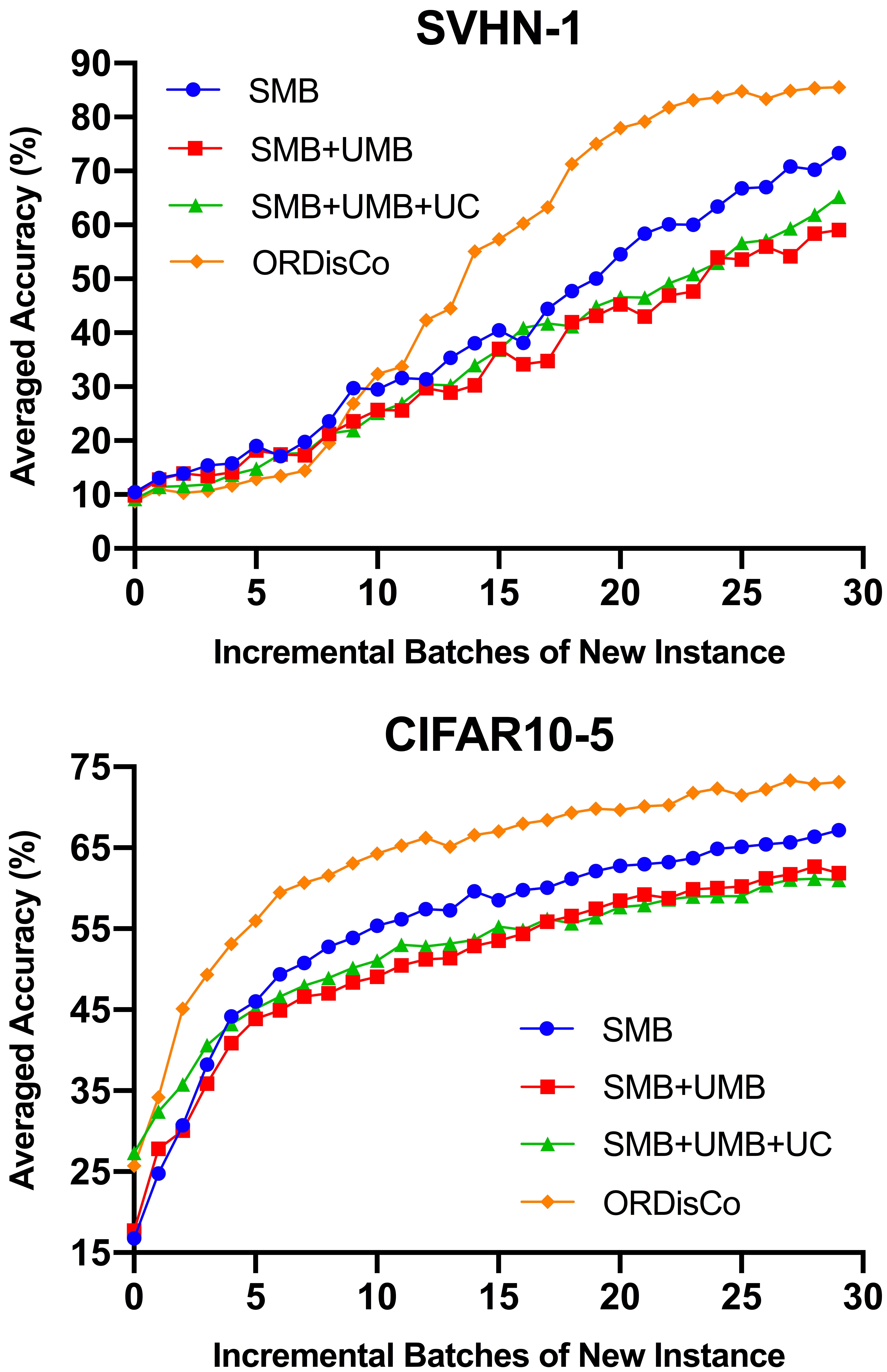}
    \caption{Averaged accuracy (\(\%\)) of SSCL on SVHN-1 and CIFAR10-5. ORDisCo (Ours) improves SMB by \(12.20\%\) and \(5.95\%\) at the final batch on SVHN-1 and CIFAR10-5, respectively. UMB: An unsupervised memory buffer of a similar size as our generator; UC: Using the unified classifier to exploit UMB.}
    \label{SSCL_limit_label}
    \vspace{-0.3cm}
\end{figure}

We report the performance of ORDisCo and other baselines during SSCL of 30 incremental batches in Fig. \ref{SSCL_limit_label} (SVHN-1 and CIFAR10-5) and Appendix C (SVHN-3 and CIFAR10-13). ORDisCo significantly outperforms other baselines, particularly when the numbers of labels are smaller, i.e. SVHN-1 and CIFAR10-5.
Compared with SMB, an additional unsupervised buffer (UMB) slightly improves the performance of SSCL on SVHN-3 and CIFAR10-13, but results in overfitting on SVHN-1 and CIFAR10-5 where the labeled data are extremely limited (See empirical results and analysis of the overfitting in Appendix A). The Unified Classifier better exploits the UMB through balancing the predictions on UMB and large amounts of unlabeled data, and thus slightly improves its performance. Also, the performance is only slightly increased by using a much larger UMB (See details in Appendix C). By contrast, ORDisCo can much better exploit the unlabeled data and substantially improve SSCL.
We then evaluate ORDisCo on Tiny-ImageNet-5 with larger number of classes and larger-scale images in Table \ref{tiny}. Compared with SMB and SMB+UMB, ORDisCo achieves a much better performance on 10-class and 20-class of size \(32 \times 32 \) and 10-class of size \(64 \times 64\), through effective usage of the incremental unlabeled data to improve classification.

Note that the reported performance is the result after training on each semi-supervised batch till convergence. Therefore, the outstanding performance of ORDisCo on limited partially labeled data, i.e. smaller numbers of labels and semi-supervised batches, indicates that ORDisCo can \emph{quickly learn a usable model} rather than waiting for more data to be collected.

\begin{table}[t]
	\centering
	\caption{Ablation study on SVHN-1 and CIFAR10-5. Both online semi-supervised generative replay (Replay) and regularization of discrimination consistency (Reg) substantially improve SSCL.}\smallskip
	\resizebox{1\columnwidth}{!}{
	\begin{tabular}{ccccc}
		\specialrule{0.01em}{1.2pt}{1.5pt}
        \multicolumn{1}{c}{}  & \multicolumn{2}{c}{SVHN-1} & \multicolumn{2}{c}{CIFAR10-5} \\
        Method & 15 Batch & 30 Batch & 15 Batch & 30 Batch \\
	   \specialrule{0.01em}{1.5pt}{1.5pt}
       SMB  &38.08  &73.32 &59.61 & 67.18 \\
	   ORDisCo (+Replay, -Reg) &31.14 &84.66 &64.38 &71.69 \\
       ORDisCo (+Replay, +Reg)  &\textbf{55.07} &\textbf{85.52} &\textbf{66.58} &\textbf{73.13} \\
      \specialrule{0.01em}{1.5pt}{1.5pt}
	\end{tabular}
	}
	\label{ablation}
	\vspace{-.2cm}
 \end{table}
 
\subsection{Ablation Study and Analysis}
Next, we analyze why ORDisCo improves continual learning of unlabeled data. ORDisCo enables a generative model to continually learn conditional generation from incremental semi-supervised data. Thus, the conditional generator of ORDisCo can much better cover the training data distribution than an unsupervised memory buffer of a similar size (Fig.~\ref{coverage}). 
We make the ablation study (SVHN-1 and CIFAR10-5 in Table \ref{ablation}, SVHN-3 and CIFAR10-13 in Appendix C) to validate the effects of the proposed strategies in ORDisCo, that both the online semi-supervised generative replay (refer to as Replay) and the regularizer to stabilize discrimination consistency (refer to as Reg) can significantly improve SSCL. The ablation study shows that the effects of the regularization on discrimination consistency are much more significant when the data sources are \emph{limited}, i.e. fewer labels and semi-supervised batches.

\begin{table}[t]
	\centering
	\caption{Averaged accuracy (\(\%\)) of a jointly trained classifier to predict conditional samples learned from earlier batches on SVHN-1 and CIFAR10-5, where regularization of discrimination consistency (Reg) significantly improves generative replay. }\smallskip
	\resizebox{1\columnwidth}{!}{
	\begin{tabular}{ccccc}
		\specialrule{0.01em}{1.2pt}{1.5pt}
        \multicolumn{1}{c}{}  & \multicolumn{2}{c}{SVHN-1} & \multicolumn{2}{c}{CIFAR10-5} \\
        Method & 5 Batch & 15 Batch & 5 Batch & 15 Batch \\
	   \specialrule{0.01em}{1.5pt}{1.5pt}
       SMB &\textbf{27.37}  &51.45  &43.37  &70.64  \\
	   ORDisCo (+Replay, -Reg)&16.95  &37.89 &49.98 &68.67  \\
       ORDisCo (+Replay, +Reg)&25.46 &\textbf{64.93} &\textbf{64.15} &\textbf{81.13}  \\
      \specialrule{0.01em}{1.5pt}{1.5pt}
	\end{tabular}
	}
	\label{generated data accuracy}
	\vspace{-.2cm}
 \end{table}

To verify how the discrimination consistency improves SSCL, we use a jointly trained classifier on the semi-supervised dataset, which achieves considerable performance of classification (\(94.38\%\) for SVHN, \(87.37\%\) for CIFAR10), to evaluate conditional samples.
We present the accuracy of the classifier to predict conditional samples on the earlier 5 and 15 batches of SVHN-1 and CIFAR10-5, which include only a few labels per batch.
As shown in Table. \ref{generated data accuracy}, the predictions on the conditional samples of ORDisCo are significantly better than the one of a sequentially trained conditional GAN with SMB. Because the quality of conditional samples is generally poor when the partially labeled data are limited, replay of such conditional samples to the classifier in ORDisCo will decrease its performance to predict pseudo-labels of unlabeled data, and further interfere with conditional generation. While, the regularizer on discrimination consistency largely alleviates this issue through stabilizing the learned distribution of unlabeled data in the discriminator, which is critical to SSCL.

\subsection{SSCL of New Class}
ORDisCo can be naturally extended to incremental learning of New Class in semi-supervised scenarios. We consider an incremental dataset consisting of 5 binary classification tasks, where the partially labeled data of two new classes are introduced in each task. We keep the total number of labels the same as the one used in New Instance. As shown in Table \ref{New Class}, ORDisCo significantly outperforms SMB, SMB+UMB and SMB+UMB+UC through more effective usage of the unlabeled data.

\begin{table}[t]
	\centering
	\caption{Averaged accuracy (\%) of SSCL of New Class. We follow the same semi-supervised splits as New Instance on SVHN-1, CIFAR10-5, or 10-class Tiny-ImageNet-5 of the size \(32 \times 32\), but construct a sequence of 5 binary classification tasks.}
	\smallskip
	\resizebox{1.00\columnwidth}{!}{
	\begin{tabular}{clccccc}
		\specialrule{0.01em}{1.2pt}{1.5pt}
	\multicolumn{1}{c}{Split}  &	\multicolumn{1}{c}{Method}  & \multicolumn{1}{c}{Task 1} & \multicolumn{1}{c}{Task 2}  & \multicolumn{1}{c}{Task 3} & \multicolumn{1}{c}{Task 4} & \multicolumn{1}{c}{Task 5}    \\
         \specialrule{0.01em}{1.5pt}{1.5pt}
      	\multirow{4}*{\tabincell{c}{SVHN-1}}
        &SMB&81.80 &50.73 &39.77 &40.41 &35.46 \\
        &SMB+UMB&\textbf{83.30} &50.53 &35.92 &33.00 &37.35 \\
        &SMB+UMB+UC&83.04 &53.10 &36.18 &36.83 &38.33 \\
        &ORDisCo&82.72 &\textbf{54.69} &\textbf{60.68} &\textbf{56.18} &\textbf{53.79} \\
        \specialrule{0.01em}{1.5pt}{1.5pt}
        \multirow{4}*{\tabincell{c}{CIFAR10-5}}
        &SMB&\textbf{95.78} &74.54 &61.63 &56.85 &55.12 \\
        &SMB+UMB&95.50 &76.68 &63.22 &62.34 &60.52 \\
        &SMB+UMB+UC&95.67 &\textbf{77.94} &65.95 &63.33 &62.43 \\
        &ORDisCo&95.60 &77.23 &\textbf{68.62} &\textbf{66.49} &\textbf{65.91}  \\
      \specialrule{0.01em}{1.5pt}{1.5pt}
        \multirow{4}*{\tabincell{c}{TinyImageNet-5}}
        &SMB&88.50  &75.17  &67.17 &67.58  &65.47  \\
        &SMB+UMB &87.83 &80.17  &68.33 &71.33 &67.13  \\
        &SMB+UMB+UC&\textbf{88.67} &79.08 &69.94 &71.13 &69.87 \\
        &ORDisCo&88.50 &\textbf{83.17} &\textbf{70.56} &\textbf{73.92} &\textbf{71.07 } \\
		\specialrule{0.01em}{1.5pt}{1.5pt}

	\end{tabular}
	}
	\label{New Class}
	\vspace{-.4cm}
 \end{table}

\subsection{Conditional Generation in SSCL}
We present the conditional samples from ORDisCo and a sequentially trained conditional GAN with SMB during SSCL in Fig. \ref{conditional generation in SSCL} (See complete conditional samples of ORDisCo in Appendix D), where conditional generation is improved from incremental learning of partially labeled data. Due to more effective usage of unlabeled data, ORDisCo can generate high-quality conditional samples in SSCL, significantly better than the SMB baseline. 
 
 \begin{figure}[t]
    \centering
    \subfigure[SVHN-3]{\includegraphics[height = 1.9cm]{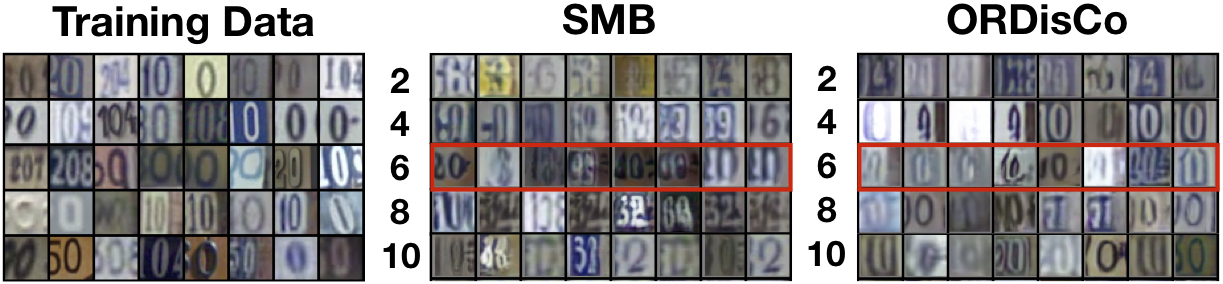}} \\
    \vspace{-.1cm}
    \subfigure[CIFAR10-13]{\includegraphics[height = 1.9cm]{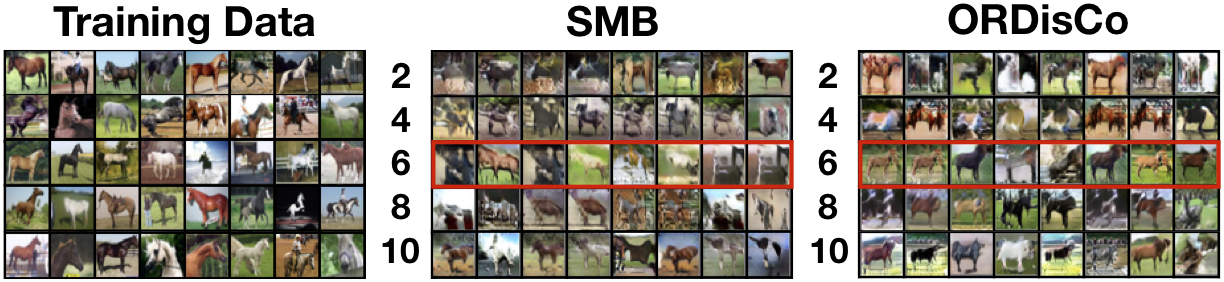}}
    \caption{Conditional generation in SSCL. We show the conditional samples of `0' in SVHN-3 and `horse' in CIFAR10-13. The column denotes the number of incremental batches learned by the conditional generator to generate images in each row. Starting from the sixth batch (the red box), ORDisCo can generate high-quality conditional samples, significantly better than the SMB baseline.}
    \label{conditional generation in SSCL}
    \vspace{-.4cm}
\end{figure}

\section{Conclusion}
In this work, we provide a systematic analysis of semi-supervised continual learning, which is a realistic yet challenging setting without extensive study. Then we propose a novel method to continually learn a conditional GAN with a classifier from partially labeled data. Extensive evaluations on various benchmarks show that our method can effectively capture and exploit the unlabeled data in both classification tasks and conditional generation. Since our method can be a plug-and-play approach for semi-supervised continual learning, a wide range of semi-supervised learning approaches for classification and conditional generation can be flexibly implemented into the framework, which is our further direction. Another further work is to use ORDisCo on Mindspore\footnote{https://www.mindspore.cn}, a new deep learning computing framework.


\section*{Acknowledgement}

This work was supported by the National Key Research and Development Program of China (Nos. 2017YFA0700904, 2020AAA0104304), NSFC Projects (Nos. 61620106010, 62076145), Beijing NSF Project (No. L172037), Tsinghua-Huawei Joint Research Program. C. Li was supported by the fellowship of China postdoctoral Science Foundation (2020M680572), and the fellowship of China national postdoctoral program for innovative talents (BX20190172) and Shuimu Tsinghua Scholar.

\bibliographystyle{plain}
\bibliography{egbib}

\clearpage

\appendix
\section{A Systematic Analysis of CL Methods}

\begin{figure}[th]
    \centering
    \includegraphics[width=1\linewidth]{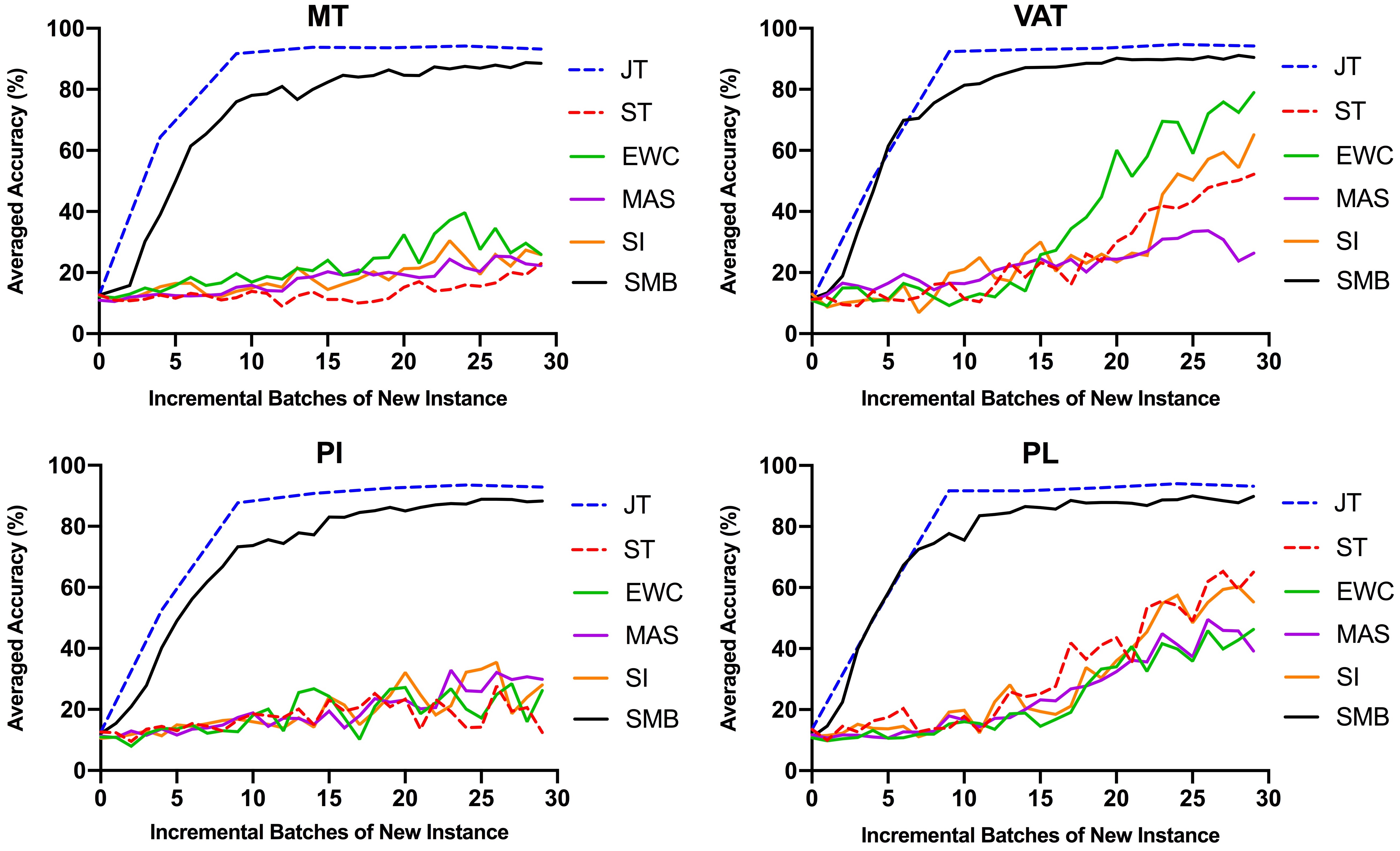}
    \caption{Baselines that combine SSL classifiers and representative CL methods to address SSCL. JT: Joint training of all the training samples ever seen. ST: Sequential training on the incremental data; SMB: ST with a supervised memory buffer to replay the labeled data ever seen.}
    \label{SSL+CL}
\end{figure}

\begin{figure}[th]
    \centering
    \includegraphics[width=1\linewidth]{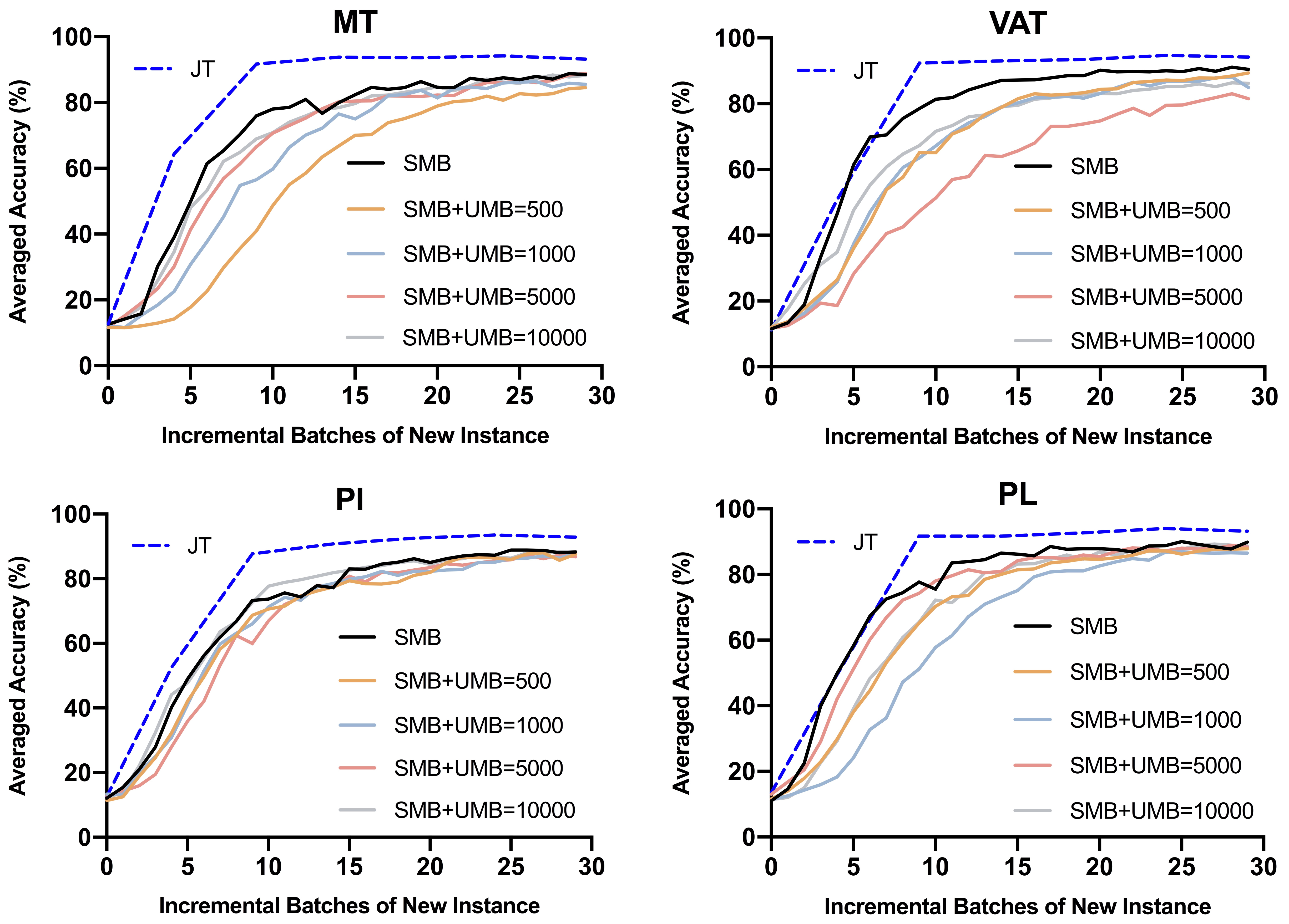}
    \caption{Performance of the baselines that combine SMB with an additional unsupervised buffer (UMB) of various sizes.}
    \label{SSL+CL_USV}
\end{figure}

In this section, we present the results of our preliminary analysis to evaluate representative continual learning (CL) strategies in semi-supervised continual learning (SSCL). As shown in Fig. \ref{SSL+CL}, for all the four semi-supervised classifiers, sequential training (ST) on the incremental partially labeled data significantly underperforms joint training (JT) of all the data ever seen, suggesting the existence of catastrophic forgetting. Also, implementation of weight regularization methods on a semi-supervised classifier cannot address this issue in SSCL. While, using a supervised memory buffer (SMB) to replay old labeled data can effectively alleviate catastrophic forgetting, but still remains a performance gap to JT, which is caused by the \emph{catastrophic forgetting of unlabeled data}. However, an additional unsupervised memory buffer (UMB) cannot effectively exploit the incremental unlabeled data to improve SSCL. We extensively evaluate the UMB of various sizes in Fig. \ref{SSL+CL_USV}. Even the large UMB with 10000 images (the size is twice of a typical generator used in ORDisCo) cannot further improve the performance from mitigating the catastrophic forgetting of unlabeled data, while a smaller UMB with 500 or 1000 images results in severe overfitting and thus interferes SSCL. Therefore, a common issue of existing CL strategies is the lack of ability to continually exploit unlabeled data in SSCL.

From the empirical analysis, we observe that using SMB can effectively alleviate catastrophic forgetting in SSCL. Classical memory replay methods for supervised CL~\cite{rebuffi2017icarl,castro2018end} use a mean-of-feature strategy to select samples, where the images that are closest to the feature mean of a class are stored in the memory buffer. A commonly-used size of the memory buffer is 20 samples per class, i.e. 200 samples for 10 classes. While, compared with supervised CL, the labeled data are generally insufficient in SSCL to perform selection, e.g., SVHN-1 only provides 300 labeled data in total. Here we adapt the mean-of-feature strategy to select samples from small amounts of labeled data (SMB-mof) in SVHN-3, and compare it with replay of all the labeled data ever seen (SMB), which is similar to unified sampling and is indeed competitive to existing selection methods as analyzed in~\cite{chaudhry2018riemannian}. We use a fixed size of 200 samples for SMB-mof, so SMB-mof is equal to SMB in the early 6 batches (180 labeled data in total) and then perform selection. The better performance of SMB than SMB-mof, particularly when more labeled data are introduced in the later batches, indicates that selecting representative labeled data via mean-of-feature still suffers from the catastrophic forgetting of labeled data. Since our motivation is to improve continual learning of unlabeled data in SSCL, we replay all the labeled data ever seen in SMB. A potential future work can be designing proper strategy to effectively select labeled data for memory buffer in SSCL.

\begin{table}[th]
	\centering
	\caption{Comparison of unified sampling (SMB) and mean-of-feature (SMB-mof) to select the supervised memory buffer.}\smallskip
	\resizebox{1.00\columnwidth}{!}{
	\begin{tabular}{cccccc}
		\specialrule{0.01em}{1.2pt}{1.5pt}
        & & Batch 15 & Batch 20 & Batch 25 & Batch 30 \\
         \specialrule{0.01em}{1.5pt}{1.5pt}
      	\multirow{2}*{\tabincell{c}{MT}}
        &SMB&79.93 &86.40 &87.57 &88.55 \\
        &SMB-mof&79.95 &83.76 &84.40 &85.03 \\
		\specialrule{0.01em}{1.5pt}{1.5pt}
		\multirow{2}*{\tabincell{c}{VAT}}
        &SMB&87.14 &88.55 &90.02 &90.44 \\
        &SMB-mof&86.33 &87.20 &87.10 &88.65 \\
        \specialrule{0.01em}{1.5pt}{1.5pt}
        \multirow{2}*{\tabincell{c}{PI}}
        &SMB&77.20 &86.20 &87.30 &88.32 \\
        &SMB-mof&78.41 &83.62 &80.87 &85.67 \\
      \specialrule{0.01em}{1.5pt}{1.5pt}
        \multirow{2}*{\tabincell{c}{PL}}
        &SMB&86.54 &87.88 &88.79 &89.88 \\
        &SMB-mof&82.24 &86.26 &85.48 &86.72 \\
      \specialrule{0.01em}{1.5pt}{1.5pt}
	\end{tabular}
	}
	\label{select SMB}
 \end{table}

To visualize the data coverage of UMB and ORDisCo on the training data distribution in Fig. \ref{coverage}, we jointly train a feature embedding layer on SVHN and apply t-SNE projection to visualize the feature embedding of the unlabeled data in UMB, the generated data sampled from ORDisCo and all the training data. The size of the UMB is around \(10\%\) of all the training data, which is comparable to the storage cost of a typical generator used in ORDisCo. However, the coverage of the generated data is substantially better than the UMB with a similar size.

\section{Implementation of SSL Classifiers and CL Methods}
For all the semi-supervised classifiers, we jointly train 200,000 iterations with the same semi-supervised split as \cite{oliver2018realistic}. To validate our implementations, in Table \ref{vs realistic} we summarize the performance of our implementations and the published performance \cite{oliver2018realistic} on SVHN with 1000 labels.

Implementation details of SSL classifiers and CL methods are as follows: For sequential training, we search the number of training epochs among 100, 200, 500 and 1000. 500 training epochs within each batch gives the best average results on the four SSL methods.
As for regularization-based SSCL extension (EWC, SI and MAS), we search the best lambda among 0.1, 1, 10 and 100 for EWC and SI, and $10^{-6}$, $10^{-4}$, $10^{-2}$ and 1 for MAS, respectively. 
Since the suitable number of training epochs may change, we do an additional grid search in 250, 500 and 750 for all the three CL methods and choose the number of epochs that achieve the best performance.
The results of lambda search are summarized in Table \ref{labmda selection}. We demonstrate the best accuracy of each trial.

\begin{table}[t]
	\centering
	\caption{Implementations of SSL classifiers on SVHN dataset with 1000 labels. We present the averaged accuracy (\(\%\)) on the test set of SVHN. We follow similar implementations as \cite{oliver2018realistic} and achieve a comparable performance as the published performance. The slight variance of performance might be caused by different random seeds.}\smallskip
    \resizebox{0.95\columnwidth}{!}{
	\begin{tabular}{cccc}
		\specialrule{0.01em}{1.2pt}{1.5pt}
         &Our Implementation & Published Performance \cite{oliver2018realistic}\\
        \specialrule{0.01em}{1.5pt}{1.5pt}
        MT &93.21 &94.35 \\
        VAT &94.22 &94.37 \\
        PI &92.89 &92.81 \\
        PL &93.19 &92.38 \\
      \specialrule{0.01em}{1.5pt}{1.5pt}
	\end{tabular}
    }
	\label{vs realistic}
 \end{table}

\begin{table}[t]
	\centering
	\caption{Hyperparameter search of weight regularization methods in SSL classifiers to address SSCL on SVHN-3. We present the best accuracy (\(\%\)) during incremental learning of 30 partially labeled batches.}\smallskip
 	\resizebox{0.90\columnwidth}{!}{
	\begin{tabular}{cccccc}
		\specialrule{0.01em}{1.2pt}{1.5pt}
        & lambda & MT & PI & PL & VAT \\
         \specialrule{0.01em}{1.5pt}{1.5pt}
      	\multirow{4}*{\tabincell{c}{EWC}}
        &0.1 &21.23 &29.61 &16.84 &66.97 \\
        &1 &45.76 &28.15 &33.50 &61.51 \\
        &10 &23.54 &28.25 &18.56 &78.96 \\
        &100 &35.80 &28.45 &18.28 &29.61 \\
		\specialrule{0.01em}{1.5pt}{1.5pt}
		\multirow{4}*{\tabincell{c}{SI}}
        &0.1 &40.36 &35.45 &60.28 &20.38 \\
        &1 &32.22 &27.56 &47.64 &65.13 \\
        &10 &30.04 &31.36 &45.92 &40.65 \\
        &100 &16.21 &27.37 &64.64 &19.54 \\
        \specialrule{0.01em}{1.5pt}{1.5pt}
        \multirow{4}*{\tabincell{c}{MAS}}
        &$10^{-6}$ &36.95 &32.90 &49.58 &33.70 \\
        &$10^{-4}$ &23.52 &21.41 &20.20 &15.87 \\
        &$10^{-2}$ &20.46 &17.78 &15.84 &13.51 \\
        &1 &12.66 &12.15 &11.47 &11.75 \\
      \specialrule{0.01em}{1.5pt}{1.5pt}
	\end{tabular}
 	}
	\label{labmda selection}
 \end{table}

\section{SSCL of New Instance on SVHN-3 and CIFAR10-13}

\begin{figure}[ht]
    \centering
    \includegraphics[width=0.80\linewidth]{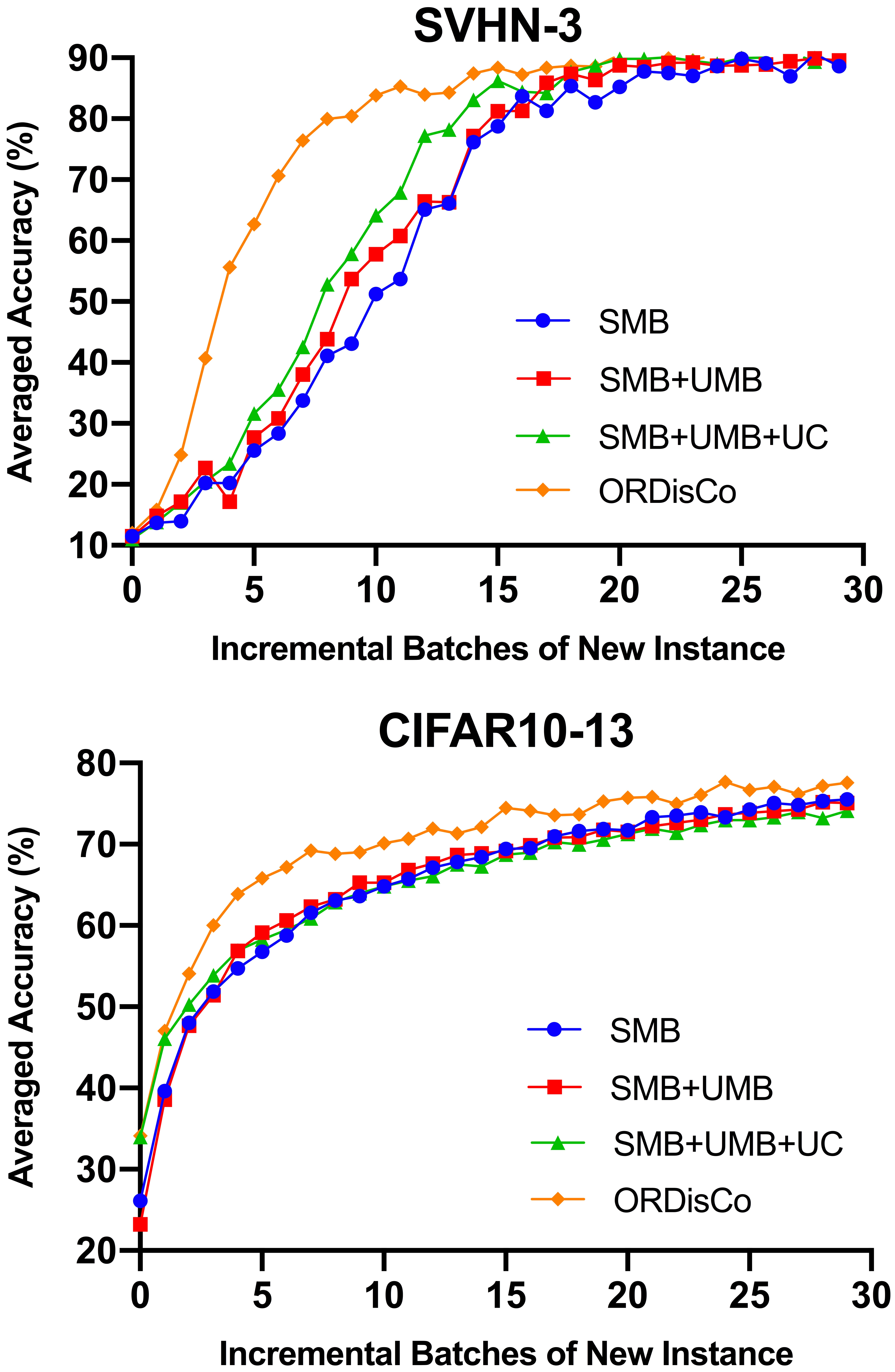}
    \caption{Averaged accuracy (\(\%\)) of classification on CIFAR10. UMB: An unsupervised memory buffer of a similar size as our generator; UC: Using the unified classifier to exploit UMB.}
    \label{SSCL_more_label}
\end{figure}

The performance of SSCL of new instance on SVHN-3 and CIFAR10-13 is summarized in Fig. \ref{SSCL_more_label}, with ablation study in Table \ref{ablation_more_labels}. ORDisCo achieves a much better performance than other baselines, particularly on fewer incremental batches of partially labeled data. The improved performance on smaller amounts of data indicates that ORDisCo can \emph{quickly learn a usable model} from incremental data, without waiting for more data to be collected. The ablation study also shows that the online semi-supervised generative replay substantially improves SSCL, while the regularization of discrimination consistency is more effective when the partially labeled data are limited, i.e. fewer incremental batches (Table \ref{ablation_more_labels}) and labels (Table \ref{ablation}).

\begin{table}[t]
	\centering
	\caption{Ablation study on SVHN-3 and CIFAR10-13. The online semi-supervised generative replay (Replay) substantially improves SSCL, while the regularization of discrimination consistency (Reg) is more effective on fewer incremental batches (15 Batch).}\smallskip
	\resizebox{1\columnwidth}{!}{
	\begin{tabular}{ccccc}
		\specialrule{0.01em}{1.2pt}{1.5pt}
       \multicolumn{1}{c}{}  & \multicolumn{2}{c}{SVHN-3} & \multicolumn{2}{c}{CIFAR10-13} \\
        Method & 15 Batch & 30 Batch & 15 Batch & 30 Batch \\
	   \specialrule{0.01em}{1.5pt}{1.5pt}
       SMB &76.16 &88.60 &64.23 &74.22 \\
	   ORDisCo (+Replay, -Reg) &85.44 &89.03 &71.70 &\textbf{78.59} \\
       ORDisCo (+Replay, +Reg) &\textbf{87.44} &\textbf{90.75} &\textbf{74.45} &77.56 \\
      \specialrule{0.01em}{1.5pt}{1.5pt}
	\end{tabular}
	}
	\label{ablation_more_labels}
	\vspace{-.1cm}
 \end{table}
 
Next, we consider to use a much larger UMB. The original size of UMB (around 5000 images) is comparable to the one of a typical generator used in ORDisCo. While, the larger UMB (around 20000 images) is 4 times larger than the original one and also much larger than all the memory cost of both the generator and the discriminator to train ORDisCo. As shown in Table \ref{larger umb}, to much enlarge the UMB with or without the unified classifier (UC) \cite{hou2019learning} can only slightly improve SSCL but substantially underperforms ORDisCo. 

 \begin{table}[th]
	\centering
	\caption{SMB with larger UMB. To use a larger UMB (4\(\times\)UMB) with or without the unified classifier (UC) only slightly improves SSCL of new instance, but significantly underperforms ORDisCo.}\smallskip
	\resizebox{1\columnwidth}{!}{
	\begin{tabular}{ccccc}
		\specialrule{0.01em}{1.2pt}{1.5pt}
        \multicolumn{1}{c}{}  & \multicolumn{2}{c}{SVHN-1} & \multicolumn{2}{c}{CIFAR10-5} \\
        Method & 15 Batch & 30 Batch & 15 Batch & 30 Batch \\
	   \specialrule{0.01em}{1.5pt}{1.5pt}
       SMB &38.08  &73.32 &59.61 & 67.18 \\
       SMB+UMB &30.26 &59.06 &52.90 &61.88 \\
       SMB+UMB+UC &33.99 &65.23 &53.61 &61.03 \\
       SMB+4\(\times\)UMB &30.74 &62.96 &57.94 &65.45 \\
       SMB+4\(\times\)UMB+UC &31.50 &66.93 &58.45 &66.26 \\
       ORDisCo &\textbf{55.07} &\textbf{85.52} &\textbf{66.58} &\textbf{73.13} \\
      \specialrule{0.01em}{1.5pt}{1.5pt}
      \\
      \specialrule{0.01em}{1.5pt}{1.5pt}
       \multicolumn{1}{c}{}  & \multicolumn{2}{c}{SVHN-3} & \multicolumn{2}{c}{CIFAR10-13} \\
        Method & 15 Batch & 30 Batch & 15 Batch & 30 Batch \\
	   \specialrule{0.01em}{1.5pt}{1.5pt}
       SMB  &76.16 &88.60 &64.23 &74.22 \\
       SMB+UMB  &77.14 &89.54 &68.89 &75.06 \\
       SMB+UMB+UC &83.11 &89.36 &67.26 &74.12 \\
       SMB+4\(\times\)UMB &79.72 &90.20 &69.42 &75.96 \\
       SMB+4\(\times\)UMB+UC &80.55 &89.05 &69.03 &75.56 \\
       ORDisCo &\textbf{87.44} &\textbf{90.75} &\textbf{71.35} &\textbf{77.56}\\
      \specialrule{0.01em}{1.5pt}{1.5pt}
	\end{tabular}
	}
	\label{larger umb}
 \end{table}

\section{Conditional Samples}
\begin{figure}[th]
    \centering
    \subfigure[SVHN-3]{\includegraphics[height = 5.0cm]{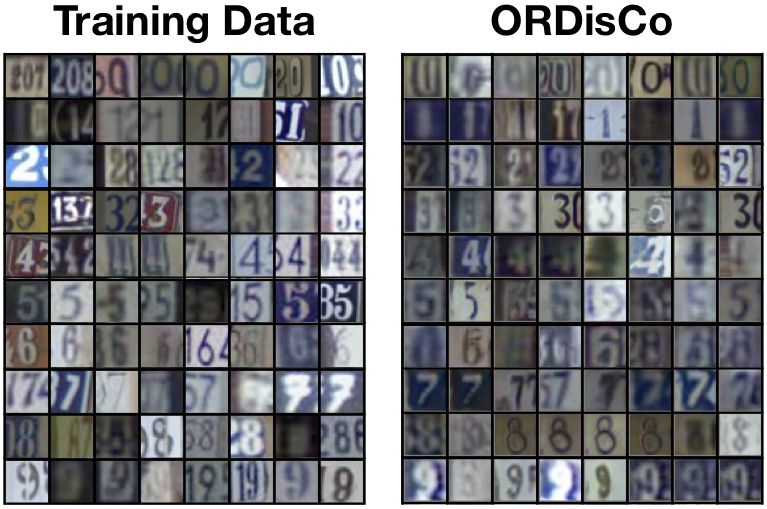}} \\
    \subfigure[CIFAR10-13]{\includegraphics[height = 5.0cm]{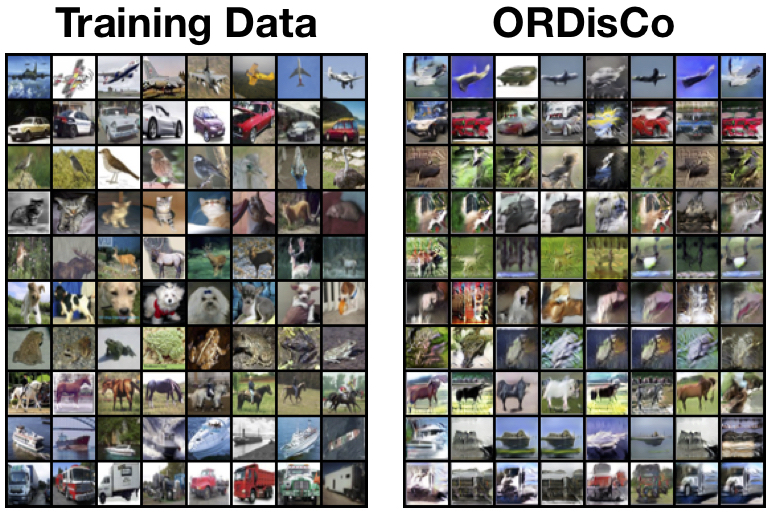}}
    \caption{Conditional generation in SSCL. We show the conditional 
    samples of ORDisCo after incremental learning of 10 batches on SVHN-3 (a) and CIFAR10-13 (b).}
    \label{conditional generation in SSCL Appendix}
\end{figure}

Here we show the conditional samples of ORDisCo learned from incremental partially labeled data in Fig. \ref{conditional generation in SSCL Appendix}.

\section{Complexity Analysis of Generative Replay}

\begin{figure}[t]
    \centering
    \includegraphics[width=0.95\linewidth]{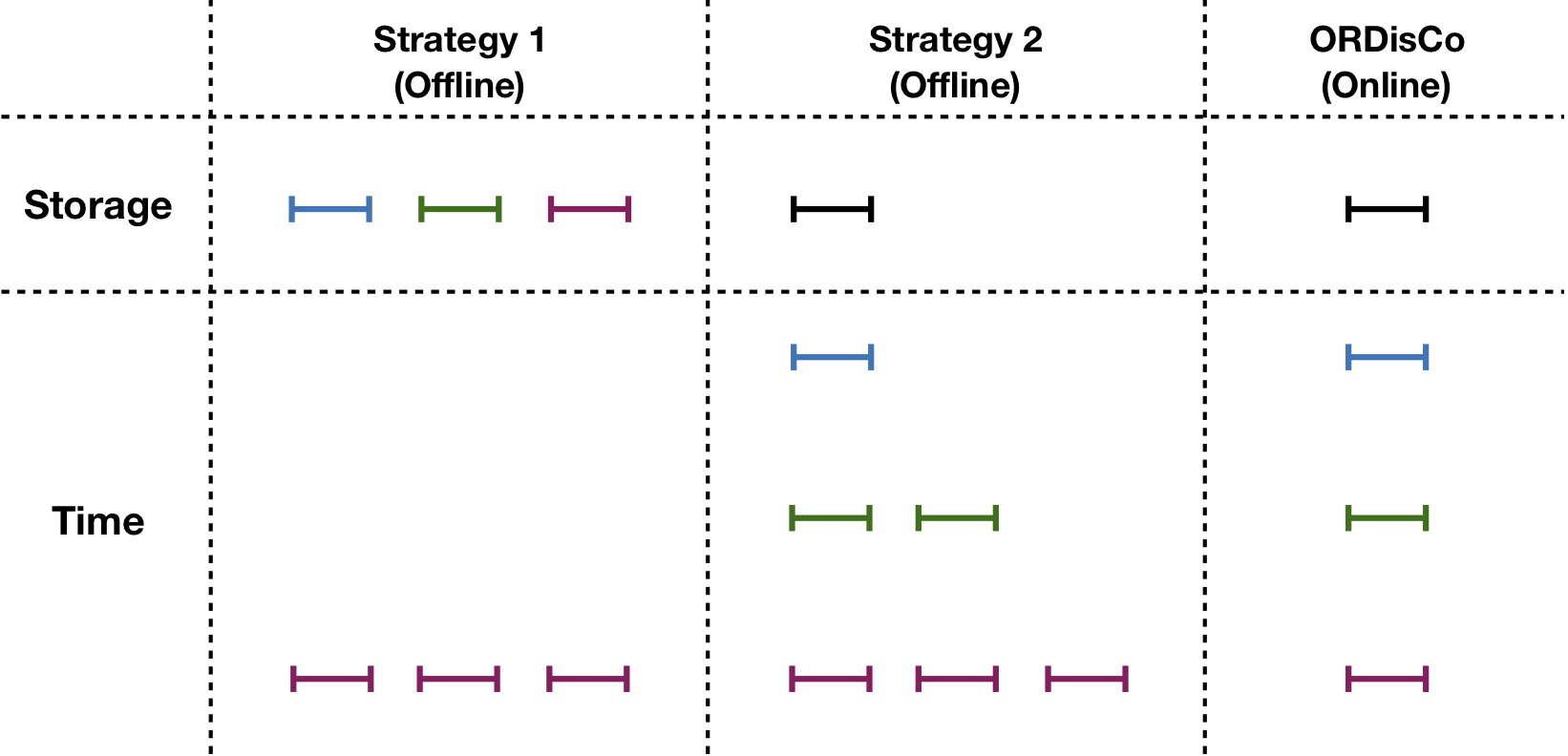}
    \caption{Storage and time complexity analysis of two offline strategies and our online strategy. We use 3 batches as an example to illustrate the storage and time complexity of the three strategies. The blue, green, and purple lines represent the storage and time costs of the first, the second and the third batch, respectively. While the black line represents the model is continually updated.}
    \label{Complexity Analysis}
\end{figure}

We propose an online semi-supervised generative replay strategy in ORDisCo, which is a time- and storage-efficient way to exploit the incremental data. We provide a complexity analysis in Table \ref{complexity} and Fig. \ref{Complexity Analysis}, of two commonly-used strategies that sample and replay generated data in an offline manner, and our online strategy. Generative replay applies generative models to continually recover the learned data distribution to overcome catastrophic forgetting. Many existing work replay generated data in an offline manner, which can be conceptually separated as two strategies: (1) All the generators learned on each task or batch are saved, and replay conditional samples to a classifier for inference \cite{ostapenko2019learning}; and (2) After training on each task or batch, the generator is saved to replay conditional samples with the next task or batch to learn a new generator. Then the saved generator is updated by the new one \cite{shin2017continual,wu2018memory}. 

\begin{figure}[t]
    \centering
    \includegraphics[width=0.95\linewidth]{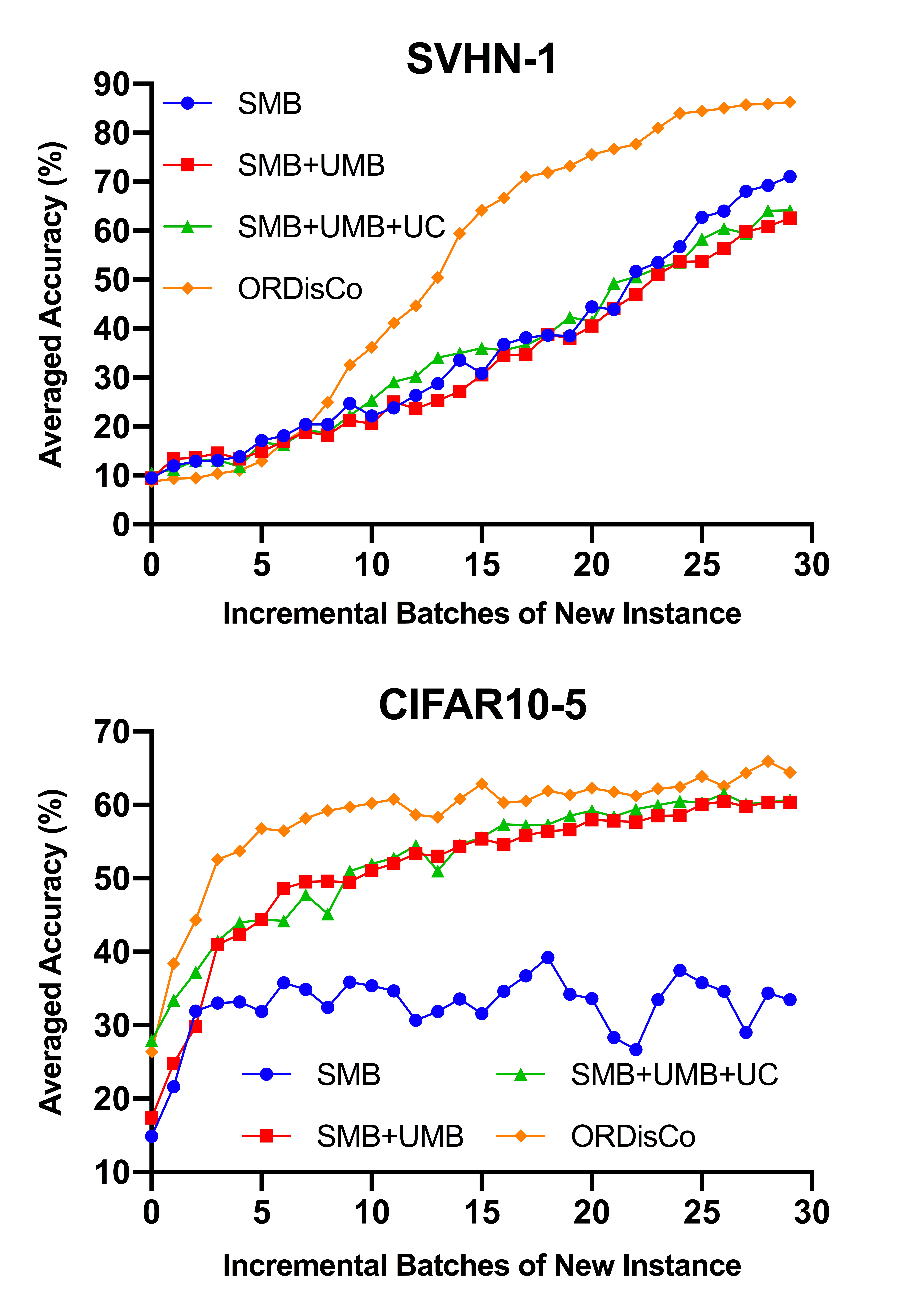}
    \caption{Selecting SMB of a fixed size through the mean-of-feature approach. SMB: A supervised memory buffer of a fixed size, selected by the mean-of-feature approach; UMB: An unsupervised memory buffer of a similar size as our generator; UC: Using the unified classifier to exploit UMB.}
    \label{SMB-mof}
\end{figure}

As shown in Fig. \ref{Complexity Analysis}, Strategy 1 has to store the learned data distribution of each task or batch as a dedicated generator and replays them from each generator for inference. Thus, for SSCL of \(B\) incremental batches, the storage and time complexities of Strategy 1 are O(\(B\)) and O(\(B\)), respective. Strategy 2 continually updates one generator after learning each task or batch. While, to overcome catastrophic forgetting in the generative model, Strategy 2 has to sample sufficient generated data from the old generator after learning each task for replay, even though the inference for classification is not required at that time. So the storage and time complexities of Strategy 2 are O(1) and O(\(B^2\)). If inference is required after learning each task or batch, the time complexity of Strategy 1 will be the same as Strategy 2, i.e. O(\(B^2\)). In contrast, the online replay strategy in ORDisCo continually updates one generator and replays a constant amount of generated data during training, where the storage and time complexities are O(1) and O(\(B\)). Note that although our online replay strategy has the same storage complexity as Strategy 2, the extra generator used in Strategy 2 results in more storage costs than ours.

\section{Selecting SMB of a Fixed Size}

Here we consider selecting SMB of a fixed size through a widely-used mean-of-feature approach \cite{rebuffi2017icarl,castro2018end,chaudhry2018riemannian} for ORDisCo and all the baselines. We keep SMB of the size 200 images, i.e. 20 images per class. As shown in Fig. \ref{SMB-mof}, the smaller and fixed SMB results in more severe overfitting in SSCL, while ORDisCo substantially outperforms other baselines due to more effectively exploiting the incremental unlabeled data.

\section{Hyperparameters of ORDisCo}

\(\alpha\) and \(\lambda\) are the two hyperparameters used in ORDisCo.
Following \cite{li2019triple}, we keep \(\alpha=0.5\) to balance the discriminator predictions of fake data-label pairs from generator and  classifier. While, \(\lambda\) is the hyperparameter to address catastrophic forgetting of unlabeled data in SSCL. We make a grid search of \(\lambda\) among 0.1, 0.01 and 0.001 based on training error, and keep \(\lambda=0.001\).

\end{document}